\pdfoutput=1
\documentclass{article}
\usepackage[nonatbib,final]{neurips_2021}
\usepackage{graphicx} %
\usepackage[numbers]{natbib}
\usepackage{caption} %
\usepackage{algorithm}
\usepackage{listings}
\usepackage{algorithmic}
\usepackage{amsmath}
\usepackage{booktabs}
\usepackage{multirow}
\usepackage[outdir=./]{epstopdf}
\usepackage{enumitem}
\usepackage{caption}
\usepackage{subcaption}
\usepackage{subfloat}
\usepackage{newfloat}
\usepackage{svg} %
\usepackage[normalem]{ulem}
\usepackage{framed}
\usepackage{mdframed}
\usepackage{xcolor}
\usepackage{lipsum}
\usepackage{float}
\usepackage{hyperref}
\usepackage{url}
\usepackage{amsfonts}
\usepackage{makecell}
\usepackage{array}
\usepackage{multirow}
\usepackage{dirtytalk}

\definecolor{shadecolor}{gray}{0.9}

\newlist{todolist}{itemize}{2}
\setlist[todolist]{label=$\square$}
\usepackage{pifont}

\usepackage{array}
\newcolumntype{L}[1]{>{\raggedright\let\newline\\\arraybackslash\hspace{0pt}}m{#1}}
\newcolumntype{C}[1]{>{\centering\let\newline  \\\arraybackslash\hspace{0pt}}m{#1}}
\newcolumntype{R}[1]{>{\raggedleft\let\newline \\\arraybackslash\hspace{0pt}}m{#1}}

\AtBeginDocument{%
  \providecommand\BibTeX{{%
    \normalfont B\kern-0.5em{\scshape i\kern-0.25em b}\kern-0.8em\TeX}}
}
\begin{document}

\title{Towards Large Reasoning Models: A Survey of Reinforced Reasoning with Large Language Models}

\href{}{\author{%
\textbf{Fengli Xu}$^{1}$ \quad \textbf{Qianyue Hao}$^{1}$ \quad \textbf{Zefang Zong}$^{1}$ \quad \textbf{Jingwei Wang}$^{1}$ \quad \textbf{Yunke Zhang}$^{1}$ \\ 
\textbf{Jingyi Wang}$^{1}$ \quad \textbf{Xiaochong Lan}$^{1}$ \quad \textbf{Jiahui Gong}$^{1}$ \quad 
\textbf{Tianjian Ouyang}$^{1}$ \quad \textbf{Fanjin Meng}$^{1}$ \\
\textbf{Chenyang Shao}$^{1}$ \quad \textbf{Yuwei Yan}$^{2}$ \quad \textbf{Qinglong Yang}$^{1}$ \quad  \textbf{Yiwen Song}$^{1}$ \quad \textbf{Sijian Ren}$^{1}$  \\ \textbf{Xinyuan Hu}$^{3}$ \quad \textbf{Yu Li}$^{1}$  \quad \textbf{Jie Feng}$^{1}$ \quad \textbf{Chen Gao}$^{1}$ \quad \textbf{Yong Li}$^{1}$ \\
All authors contributed equally.\\
${1}$ Tsinghua University, Beijing, China;
${2}$ HKUST (GZ), Guangzhou, China \\
${3}$ Emory University, Atlanta GA, USA \\
\texttt{fenglixu@tsinghua.edu.cn, liyong07@tsinghua.edu.cn}}}

\maketitle
\begin{abstract}
Language has long been conceived as an essential tool for human reasoning. The breakthrough of Large Language Models (LLMs) has sparked significant research interest in leveraging these models to tackle complex reasoning tasks. Researchers have moved beyond simple autoregressive token generation by introducing the concept of ``thought’’---a sequence of tokens representing intermediate steps in the reasoning process. This innovative paradigm enables LLMs' to mimic complex human reasoning processes, such as tree search and reflective thinking. Recently, an emerging trend of learning to reason has applied reinforcement learning (RL) to train LLMs to master reasoning processes. This approach enables the automatic generation of high-quality reasoning trajectories through trial-and-error search algorithms, significantly expanding LLMs' reasoning capacity by providing substantially more training data. Furthermore, recent studies demonstrate that encouraging LLMs to ``think’’ with more tokens during test-time inference can further significantly boost reasoning accuracy. Therefore, the train-time and test-time scaling combined to show a new research frontier---a path toward Large Reasoning Model. The introduction of OpenAI’s o1 series marks a significant milestone in this research direction. In this survey, we present a comprehensive review of recent progress in LLM reasoning. We begin by introducing the foundational background of LLMs and then explore the key technical components driving the development of large reasoning models, with a focus on automated data construction, learning-to-reason techniques, and test-time scaling. We also analyze popular open-source projects at building large reasoning models, and conclude with open challenges and future research directions. 
\end{abstract}

\maketitle

\section{Introduction}
\label{sec::introduction}

\say{If there is a severe deficit of language, there will be severe deficit of thought} --- Noam Chomsky

Fueled by advancements in deep learning and the availability of web-scale datasets, Large Language Models (LLMs) have emerged as a transformative paradigm on the path toward Artificial General Intelligence (AGI). These massive AI models typically adopt Transformer architecture and are pre-trained on large-scale text corpus with the next-token prediction task~\cite{zhao2023survey}. The neural scaling law demonstrates that their performance improves significantly as the model size and training data increase~\cite{kaplan2020scaling}. More importantly, LLMs also unlock remarkable \emph{emergent abilities} that are absent in smaller models~\cite{wei2022emergent}, such as in-context learning~\cite{dong2022survey}, role playing~\cite{shanahan2023role} and analogical reasoning~\cite{webb2023emergent}. These abilities allow LLMs go beyond natural language processing problems to facilitate a wider range of tasks, such as code generation~\cite{gehring2024rlef}, robotic control~\cite{ahn2022can}, and autonomous agents~\cite{deng2024mind2web}. 

Among these abilities, human-like reasoning has garnered significant attention from both academia and industry, since it demonstrates great potential for LLMs to generalize to complex real-world problems through abstract and logical reasoning. A notable breakthrough in this area is the ``chain-of-thought’’ prompting technique~\cite{wei2022chain}, which can elicit step-by-step human-like reasoning processes at test time without any additional training. Such intuitive prompting techniques have been proven effective to substantially improve the reasoning accuracy of pre-trained LLMs, which also leads to the development of more advanced prompting techniques like ``tree-of-thought’’~\cite{yao2024tree}. These approaches introduce the concept of ``thought’’ as a sequence of tokens that represents the intermediate steps in human-like reasoning process. By incorporating such intermediary steps, LLM reasoning moves beyond simple autoregressive token generation, enabling more sophisticated cognitive architectures like tree search~\cite{yao2024tree} and reflective reasoning~\cite{zeng2024perceive}. 

Recently, there has been a significant research trend in learning to reason~\cite{openai2024learning}, which seeks to train LLMs to master human-like reasoning processes. A key challenge in this research direction is the lack of training data. Human annotation is often prohibitively expensive, particularly for step-by-step reasoning trajectories that have proven effective in supervising LLM reasoning~\cite{lightman2023let}. To address this issue, recent studies have shifted from human annotation to LLM-driven search algorithms. These approaches utilize external verification for reasoning problems to automatically generate accurate reasoning trajectories through trial-and-error search~\cite{luo2024improve}. More importantly, researchers have proposed to train Process Reward Models (PRMs) on these reasoning trajectories~\cite{zhang2024rest}. PRMs can provide dense, step-wise rewards that facilitate reinforcement learning for LLM reasoning. These methods combined to reduce the reliance on human annotation data, and create a ``reinforced cycle’’ for augmenting LLM reasoning that effectively integrates ``search’’ and ``learning’’, which are the two methods that can scale endlessly as predicted by Richard Sutton~\cite{sutton2019bitter}. Therefore, this novel paradigm enables the scaling of LLM reasoning capabilities with increased train-time compute, paving the way for more advanced reasoning models.

Moreover, recent study shows scaling up test-time compute can also improve LLM reasoning accuracy. Specifically, PRMs can be used to guide LLMs to evaluate and search through the intermediate ``thoughts’’~\cite{snell2024scaling}, which encourages LLMs to generate deliberate reasoning steps during test-time computation and boosts reasoning accuracy. This approach gives rise to the test-time scaling law, which predicts spending more tokens for deliberate reasoning at test-time can improve accuracy~\cite{openai2024learning}. Therefore, the RL-driven train-time scaling and search-based test-time scaling combined to show a promising research direction to fully unleash the reasoning capabilities of LLMs, \emph{i.e.}, a path toward \emph{Large Reasoning Models}. A key milestone in this research direction is OpenAI’s o1 series~\cite{zhong2024evaluation}, which demonstrates the effectiveness of this approach and echoes OpenAI’s vision of transitioning LLMs from \emph{conversational AI} (level 1) to more powerful \emph{reasoning AI} (level 2) in the five-steps road map toward AGI~\cite{duenas2024path}. Several open-source projects, such as OpenR~\cite{wang2024openr}, LLaMA-Berry~\cite{zhang2024llama} and Journey Learning~\cite{qin2024o1}, are dedicated to reproduce the strong reasoning capacity of OpenAI's o1, providing valuable insights for developing large reasoning models.  

In this survey, we provide a comprehensive review of recent research efforts in the progression toward large reasoning models. Section~\ref{sec::background} offers a brief introduction to the background of LLM reasoning. The subsequent three sections delve into the key technical components driving the development of large reasoning models. Specifically, Section~\ref{sec::data} focuses on training data construction, emphasizing the shift from human annotation to LLM-driven automated search. Section~\ref{sec::train} reviews reinforcement learning methods that are pivotal for scaling LLM reasoning capabilities with increased train-time compute, while Section~\ref{sec::test} discusses test-time scaling with a particular emphasis on PRM-guided search. In Section~\ref{sec::implementation}, we analyze the development of OpenAI's o1 series and other open-source projects, exploring the path toward large reasoning models. Section~\ref{sec::inference} summarizes additional test-time enhancement techniques, and Section~\ref{sec::benchmark} reviews reasoning benchmarks. Finally, we conclude the survey with a discussion of open problems and future research directions.

\section{Background}
\label{sec::background}

\subsection{Pre-trianing}
As the foundational stage of training LLMs, effective pretraining is crucial for developing reasoning abilities. Before discussing pretraining for LLMs' reasoning, we first outline the basic process of general LLM pretraining. Through pretraining, LLMs not only acquire core linguistic knowledge but also gain diverse world knowledge, establishing a robust groundwork for the emergence of advanced capabilities and effective value alignment~\cite{zhao2023survey}. Typically, LLM pretraining relies on high-quality text corpora~\cite{dubey2024llama, yang2024qwen2}, including extensive collections of web content, books, codes and other types of data. Leveraging these rich textual corpora, LLMs are built on the transformer architecture that are trained with next-token prediction task. After pretraining, LLMs generally demonstrate exceptional in-context learning capabilities~\cite{brown2020language}, enabling them to generate coherent text and provide accurate answers to a wide range of questions by utilizing their vast knowledge base. Notably, the pretraining stage plays a pivotal role in cultivating the reasoning abilities of LLMs. For example, research~\cite{wei2022chain} has shown that datasets rich in code and mathematical content serve as a key foundation for developing robust reasoning skills. Following this observations, newly developed LLMs~\cite{abdin2024phi} begin to introduce carefully designed synthetic data for enhancing the reasoning abilities of LLMs. During pretraining, a critical challenge lies in balancing the proportion of code and mathematical data with general text corpora to maintain strong general linguistic abilities while unlocking the reasoning potential of LLMs.

\subsection{Fine-tuning}
While pretraining enables LLMs to exhibit reasoning abilities through in-context learning, fine-tuning techniques are widely employed to achieve zero-shot and improved reasoning capabilities for LLMs. Here, we first outline the basic fine-tuning process and then explore its potential for enhancing reasoning abilities. As described in~\cite{ouyang2022training}, after the pretraining stage, LLMs enter a supervised fine-tuning phase (SFT), also referred to as the instruction tuning stage. The primary goal of this phase is to refine the model's output style, ensuring its responses are aligned with human needs and real-life applications. This is achieved by training via diverse instruction datasets that reflect a wide range of everyday human interactions, typically created through extensive and carefully curated manual annotation and refinement~\cite{zhou2024lima}. With the advent of ChatGPT, new methods have emerged for generating diverse instruction datasets. These include techniques that distill data directly from powerful LLMs~\cite{wang2022self,xu2023wizardlm} and automated approaches for large scale dataset construction from existing corpora~\cite{wei2021finetuned, ding2023enhancing}. Using these well-crafted instruction-tuning datasets, the fine-tuning process continually uses the next-token prediction objective, similar to pretraining. However, unlike pretraining, fine-tuning specifically calculates the loss for the answers while generally ignoring the loss for the questions. Additionally, incorporating datasets that include chain-of-thought (CoT)~\cite{wei2022chain} reasoning and mathematical problem-solving examples has been shown to significantly enhance the reasoning capabilities of LLMs, making this an area of active research. Following the practice in general aspects, most current approaches leverage data distillation from advanced large reasoning models, followed by fine-tuning to enhance the reasoning capabilities of LLMs to obtain the final large reasoning models.

\subsection{Alignment}
Relying solely on direct data distillation from advanced large reasoning models limits the potential of new LLMs. A more promising approach is to use reinforcement learning for data construction and model training, which precisely corresponds to the final alignment stage in general LLM training. In the general training of LLM, the alignment phase typically involves methods such as Reinforcement Learning from Human Feedback (RLHF)~\cite{ouyang2022training} to guide the model toward generating content that meets the criteria of being helpful, harmless, and honest. The goal of this phase is to enhance the safety and controllability of LLMs in the reality. Compared to the former SFT phase, this stage usually incorporates a large amount of carefully curated, manually labeled ranking data to accurately reflect human preferences~\cite{dubey2024llama, yang2024qwen2}. This data includes not only correct demonstrations but also undesirable cases that should be avoided. Standard RLHF typically involves an SFT model, a reward model, and a aligned model, which are iteratively optimized using methods like PPO~\cite{schulman2017proximal}. Due to the high data requirements and training costs of standard RLHF, methods like Direct Preference Optimization (DPO)~\cite{rafailov2024direct}, have been proposed to reduce reliance on explicit reward models. In DPO, preference loss is defined as a function of the policy to directly guide model optimization. Given the multi-step nature and complexity of reasoning problems, alignment-based post-training has become the final and most critical step in stimulating the reasoning capabilities of LLMs. By carefully decomposing the reasoning process and gradually feeding signals back to the model, various self-training methods~\cite{gulcehre2023reinforced, kumar2024training, zhang2024rest} based on reinforcement learning and preference learning have achieved notable success.

\subsection{Prompting LLMs for Advanced Reasoning}

Human-like reasoning is one of the most important abilities that emerge in LLMs with sufficiently large model parameters~\cite{webb2023emergent}. While zero-shot reasoning may remain unreliable for some tasks, researchers have discovered various prompting techniques to enhance these capabilities. These techniques can be broadly categorized into three main approaches: step-by-step reasoning, multi-path exploration, and decomposition-based methods.

The step-by-step reasoning approach, exemplified by Chain-of-Thought prompting~\cite{wei2022chain}, demonstrates that explicitly showing intermediate reasoning steps significantly improves problem-solving abilities. Even simple prompts like ``Let's think step by step'' can effectively guide the reasoning process~\cite{kojima2022large}. This approach has been further refined through Self-Consistency~\cite{wang2022self}, which generates multiple reasoning paths to arrive at more reliable conclusions, and Auto-CoT~\cite{zhang2022automatic}, which automates the generation of effective reasoning chains.

Multi-path exploration approaches extend beyond linear reasoning by considering multiple potential solution paths simultaneously. Tree of Thoughts~\cite{yao2024tree} organizes alternative reasoning pathways in a tree structure, enabling systematic exploration of different solution strategies. Graph-of-Thoughts~\cite{besta2024graph} further generalizes this to a graph structure, allowing for more flexible reasoning patterns and backtracking capabilities. ReAct~\cite{yao2022react} enriches this paradigm by interleaving reasoning with action steps, enabling more dynamic interaction with external environments.

For complex problems, decomposition-based methods have proven particularly effective. Least-to-Most Prompting~\cite{zhou2022least} and Algorithm of Thoughts~\cite{sel2023algorithm} systematically break down complex problems into manageable components, while Plan-and-Solve~\cite{wang2023plan} provides strategic guidance for tackling these subproblems. These methods are especially valuable when dealing with tasks that require multiple steps or different levels of analysis.

These extensive reasoning capabilities, enhanced through structured prompting strategies, have proven particularly effective for tasks requiring careful analysis and systematic thinking, enabling LLMs to accomplish a wide variety of complex social scientifically relevant tasks. The success of these methods demonstrates that while LLMs possess inherent reasoning abilities, their full potential can be unlocked through careful guidance and structure in the prompting process.

\subsection{Agentic Workflow}

On top of the instruction following and in-context learning capabilities of LLMs, researchers start to design agentic workflows that program the ``thinking patterns'' of LLMs~\cite{sumers2023cognitive}. Such agentic workflows allow researchers to enhance LLM's reasoning ability without any additional training, but it often requires more test-time compute. 
In-context learning~\cite{dong2022survey,coda2024meta} is the ability to improve LLM's task specific performance by simply providing a few in-context demonstrations, enables LLMs to efficiently generalize to unseen problems without computationally expensive trainings~\cite{brown2020language}. Although the origin of such capabilities remains a largely debatable topic, recent studies suggest in-context learning improves LLMs' performance by allowing them to capture the label space, the distribution of input text, and the desired format of answers~\cite{min2022rethinking}. Such desirable features have enabled researchers to adapt general purpose LLMs to diverse task scenarios, such as simulating the perspective of certain demographic groups through in-context role play~\cite{chen2023large}. 
Recent studies suggest effective agentic workflow can largely improve LLMs abilities for simulating human behavior~\cite{park2023generative,shao2024beyond}, human-LLM interaction~\cite{maas2023infinity}, and collaborative task solving~\cite{qian2023communicative}. The ability to program LLMs with agentic workflow lays the foundation of improving LLM's reasoning capabilities with complex cognitive architecture.




\section{Data Construction: from Human Annotation to LLM Automation}
\label{sec::data}

\begin{figure}[ht]
    \centering
    \includegraphics[width=0.9\linewidth]{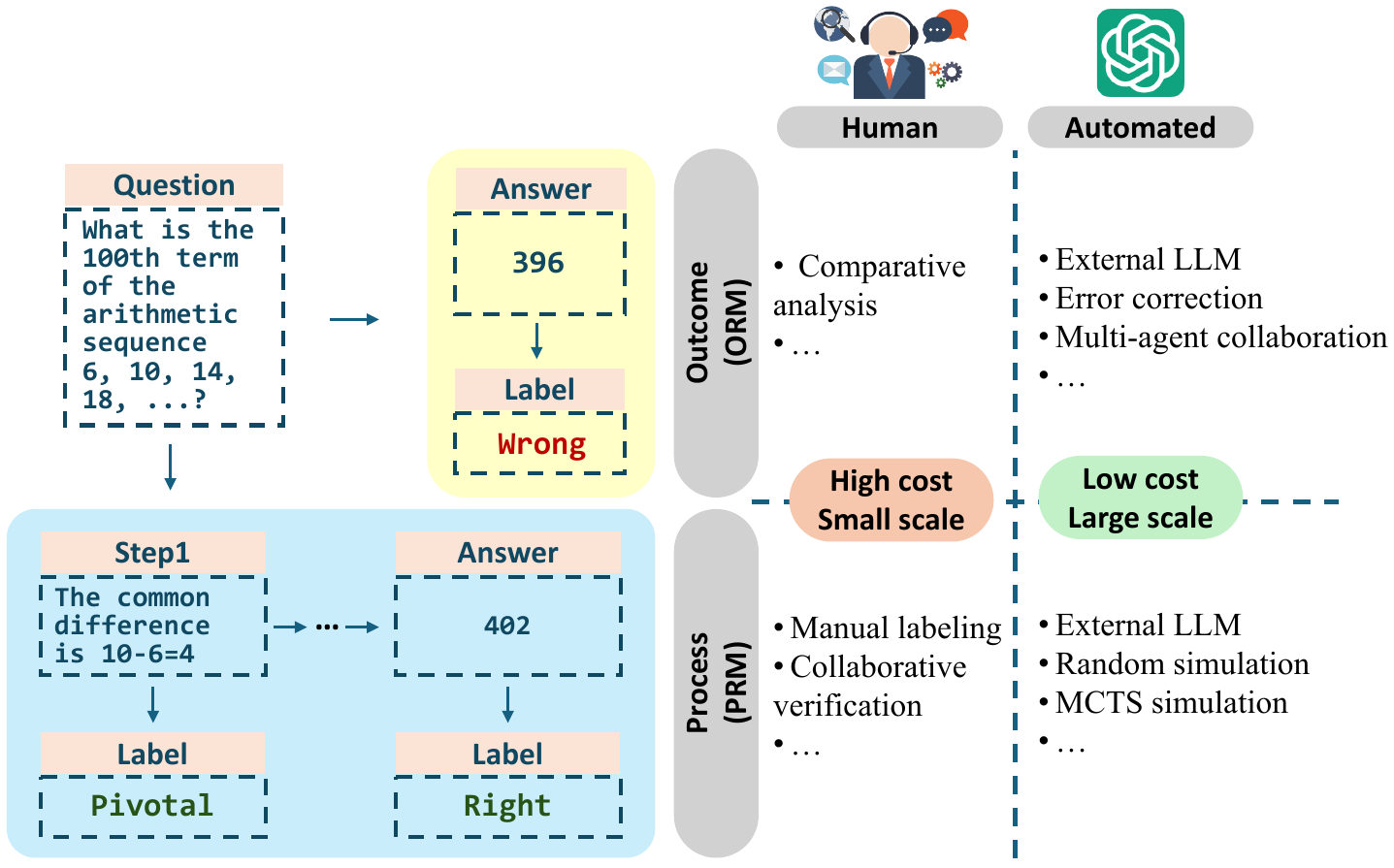}
    \caption{Illustrating different paradigms for annotating LLM reasoning data.}
    \label{fig:enter-label}
\end{figure}

Creating large-scale, high-quality reasoning datasets is crucial for enhancing the reasoning capabilities LLMs. However, this task poses significant challenges due to its high cost. As shown in Figure~\ref{fig:enter-label}, human annotation is widely considered of high-quality but is prohibitively expensive and difficult to scale. Conversely, automating the annotation process with LLMs offers a more cost-effective alternative but faces the challenge of limited validation, particularly for step-by-step reasoning processes. In this section, we review recent research efforts in this area (summarized in Table~\ref{tab:my-table}), highlighting the shift from human annotation to LLM automation.

\begin{table}[h]
\centering
\caption{Training data construction for LLM reasoning.}
\label{tab:my-table}
\resizebox{\textwidth}{!}{
\begin{tabular}{@{}cccccc@{}}
\toprule
Method                    & Label                    & Paper                    & Year                 & Task                                                                                                       & Brief Description                                                             \\ \midrule
\multirow{5}{*}{Human Annotation}      & \multirow{3}{*}{Outcome} & \cite{nasution2024chatgpt}                  &          2024   & 
\begin{tabular}[c]{@{}c@{}}Text classification \\ Semantic analysis   \end{tabular}
& Voting annotation                      
\\ \cmidrule(l){3-6} 
                            &                          & \cite{ouyang2022training}                  &     2022      &  Preference Alignment   &  Preference ranking              
                            \\ \cmidrule(l){2-6} 
                               
                            & \multirow{1}{*}{Process} & \cite{lightman2023let}   & 2023                 &  Mathematical reasoning                                                                                       & Stepwise annotation          \\ \midrule
                            
\multirow{5}{*}{\begin{tabular}[c]{@{}c@{}}Human-LLM\\Collaboration\end{tabular}}      & \multirow{5}{*}{Outcome}
                        
                    & \cite{goel2023llms}     &   2023    &  Semantic analysis        &  
                    Human correction
                        
                        \\ \cmidrule(l){3-6}   
                        
                      & & \cite{wang2024human}     &   2024    & Text classification            & Human correction
                        
                        \\ \cmidrule(l){3-6}   
                        &            & \cite{li2023coannotating}   &  2023  & \begin{tabular}[c]{@{}c@{}}Text classification\\ Semantic analysis   \end{tabular}
                                         & \begin{tabular}[c]{@{}c@{}}Task allocation\\ Uncertainty assessment\end{tabular}
                                                                               
                               \\ \midrule
\multirow{15}{*}{LLM Automation} & \multirow{5}{*}{Outcome} & \cite{puri2020training}  &  2020   &  Commonsense reasoning     
                                                                    &   Text extraction                   
                                \\ \cmidrule(l){3-6} 
                            &            & \cite{schick2024toolformer} &     2024   &   Tool use                                     &   Trial and error 
                            
                            
                                \\ \cmidrule(l){3-6} 
                            &                          & \cite{kwon2024language}    &   2024                   &   Embodied  tasks   &    Synthetic augmentation
                                \\ \cmidrule(l){3-6} 
                            &                          & \cite{qiao2024autoact}    &   2024                   &    \begin{tabular}[c]{@{}c@{}}Commonsense reasoning \\ Domain knowledge reasoning \end{tabular}  &    Multi-agent collaboration

                                                                        \\ \cmidrule(l){2-6} 
                            & \multirow{5}{*}{Process} & \cite{luo2023wizardmath} & 2023                 & Mathematical reasoning                                                                                              & Stronger LLM                                                             \\ \cmidrule(l){3-6} 
                            &                          & \cite{wang2024math}      & 2024                 & Mathematical reasoning                                                                                              & Monte Carlo simulation                                                       \\ \cmidrule(l){3-6} 
                            &                          & \cite{wang2024multi}     & 2024                 & \begin{tabular}[c]{@{}c@{}}Mathematical reasoning\\ Programming\end{tabular}                                        & Monte Carlo simulation                                                       \\ \cmidrule(l){3-6} 
                            &                          & \cite{luo2024improve}    & 2024                 & Mathematical reasoning                                                                                              & MCTS simulation                                                         
                            \\ \midrule
\multirow{5}{*}{\begin{tabular}[c]{@{}c@{}}LLM Automation\\with feedback\end{tabular}} & \multirow{3}{*}{Outcome} & \cite{lee2024llm2llm} &  2024                      &      \begin{tabular}[c]{@{}c@{}}Text classification \\ Mathematical reasoning \\ Domain knowledge reasoning   \end{tabular}                                                                 &   Self-refining
                                \\ \cmidrule(l){3-6} 
                            &                          & \cite{song2024trial}    &   2024                   &    \begin{tabular}[c]{@{}c@{}} Embodied  tasks\end{tabular}   &    Contrastive learning  
                            \\ \cmidrule(l){2-6} 
                            & Process & \cite{zhang2024rest}     & 2024                 & \begin{tabular}[c]{@{}c@{}}Mathematical reasoning\\ Domain knowledge reasoning\end{tabular} & \begin{tabular}[c]{@{}c@{}}MCTS simulation\\ Self-refining\end{tabular} \\ \bottomrule
\end{tabular}}
\end{table}

\subsection{Human Annotation} 
The role of human annotation in constructing datasets for LLMs is indispensable. Human annotators are characterized by their meticulousness, patience, and precision, as well as their adaptability to novel scenarios and capability to handle ambiguous data effectively~\cite{nasution2024chatgpt}. Zhou et al.~\cite{zhou2024lima} demonstrate that even with minimal human-annotated data, models can achieve strong performance, highlighting the critical role of carefully curated annotations in model effectiveness. Human-annotated data plays a pivotal role in enhancing the reasoning capabilities of large language models. In the context of Reinforcement Learning with Human Feedback (RLHF)\cite{ouyang2022training}, preference data from human annotators enables LLMs, initially trained on general text corpora, to align with intricate human values and complex ethical considerations. This generalizable approach of annotation helps in fine-tuning models for specific tasks. Building on this foundation, Lightman et al.\cite{lightman2023let} demonstrated the efficacy of using human annotators to evaluate the reasoning quality at each step of mathematical reasoning processes, significantly improving the accuracy of LLM reasoning. This highlights how human annotation can bridge the gap between general training data and domain-specific challenges, such as complex reasoning tasks.
\par Enhancing reasoning capabilities in LLMs requires process supervision, where human annotators guide each step of the reasoning process~\cite{lightman2023let}. However, such supervision demands extensive human-annotated data, making it resource-intensive and unsustainable. Given that LLM training typically requires terabytes of data, the volume of which is critical for model performance, constructing datasets purely through manual annotation becomes increasingly impractical. This highlights the need for alternative approaches to improve reasoning without relying solely on human annotation. One promising approach is the collaboration between humans and LLMs for annotation, where LLMs are leveraged to accelerate the process while preserving the high quality of human-generated annotations. Specifically, the annotation process can be divided into two stages: the pre-annotation stage and the refinement stage. During the pre-annotation stage, LLMs can be employed to perform an initial round of annotations, taking advantage of a few manually provided examples for a quick and efficient setup~\cite{goel2023llms, kim2024meganno+}. In the refinement stage, human annotators can assess the quality of LLM-generated annotations and focus on correcting only the subset of annotations with poor quality~\cite{kim2024meganno+, wang2024human, mikulova2023quality, goel2023llms}.
To enable scalable annotation processes, recent works have increasingly focused on how to maximize automation while ensuring data quality, thus reducing human involvement without compromising the accuracy of the annotations.

\subsection{LLM Automated Outcome Annotation} 
Data annotation is a challenging and resource-intensive task, particularly in scenarios requiring complex operations such as filtering, identifying, organizing, and reconstructing textual data. These tasks are often tedious, time-consuming, and demand significant human effort, making them a costly bottleneck in large-scale data construction efforts~\cite{tan2024large, ding2024data}. To address these challenges, leveraging LLMs for data annotation provides a cost-effective and efficient alternative. With context window lengths exceeding 100k tokens, LLMs can effortlessly process lengthy texts and large volumes of structured data~\cite{achiam2023gpt}, handling the intricate requirements of data annotation with remarkable efficiency. Their strong instruction-following capabilities~\cite{zhang2023instruction} enable them to flexibly accommodate diverse and complex annotation scenarios, while achieving a level of quality comparable to that of human annotators. By automating these demanding tasks, LLMs significantly reduce the reliance on human labor, streamlining the annotation process and enhancing overall productivity~\cite{zeng2024llmbar}.

LLMs are capable of handling a wide variety of automated annotation tasks, ranging from simple question-answer extraction~\cite{puri2020training} to the inclusion of additional target information~\cite{wei2023simple}. Without human demonstrations, LLMs rely on their powerful reasoning and in-context learning abilities to independently address more complex annotation needs. For instance, Schick et al.\cite{schick2024toolformer} demonstrated how LLMs can be used to construct datasets for tool usage. For each candidate position that may require an API call, the LLM is able to comprehend the logical relationships within the surrounding context, generate relevant questions, and identify the appropriate tool API to address the issue. 
When human demonstrations are available, LLMs can further enhance their performance by mimicking the patterns and reasoning strategies illustrated in these examples. For complex tasks, human demonstrations provide high-quality trajectories—sequences of thoughts, observations, or actions—that guide the LLMs in replicating human decision-making processes. Existing studies have shown that even zero-shot LLMs, guided by task-agnostic prompts based on human demonstrations, can perform annotation tasks effectively~\cite{kwon2024language}. Moreover, for tasks involving highly intricate and nuanced trajectories, LLMs can incorporate specialized agents, such as a Plan-Agent, Tool-Agent, and Reflect-Agent, to address different aspects of the annotation process, thereby further enhancing their ability to align with human-like reasoning and behavior~\cite{qiao2024autoact}. 
These diverse capabilities extend naturally to reasoning outcome annotation tasks, where LLMs not only infer underlying logical structures but also systematically document intermediate reasoning steps and their associated conclusions. This enables the creation of annotated datasets that capture not just final outcomes but the full reasoning processes leading to them, offering richer insights for downstream applications.

Beyond annotation with human demonstrations, LLMs can independently enhance their annotation capabilities through search with feedback, a process that involves iterative refinement by learning from dynamic environment. Failed data points can be considered a classic form of feedback, serving as valuable feedback for the model to identify weaknesses and design targeted adjustments. By self-correcting erroneous samples and generating refined training data, LLMs engage in a cycle of self-improvement that strengthens both their understanding and reasoning~\cite{lee2024llm2llm}. Furthermore, LLMs can systematically analyze the causes of their errors, extracting key insights and encoding these as self-learned knowledge to guide future reasoning tasks~\cite{li2024dotamath}. This feedback-driven approach can also involve pairing failed trajectories with successful ones based on similarity, enabling contrastive learning strategies to refine the model's parameters. Through such iterative search and refinement mechanisms, LLMs not only address errors but also develop a more robust capacity for reasoning, enabling deeper generalization and adaptability across complex tasks~\cite{song2024trial}.

\subsection{LLM Automated Process Annotation}
In complex reasoning tasks, each step of the model's output can significantly influence the final result, making it essential to label intermediate decisions as "correct," "incorrect," or assign an intermediate reward, namely process annotation.
However, manually labeling these steps is costly and time-consuming.
For example, Lightman \textit{et al.}~\cite{lightman2023let} inputs massive manual efforts to produce a large-scale process annotation dataset, i.e. PRM800K, which satisfies the requirement in training an effective process reward model (PRM) and greatly enhances the reasoning capability of LLMs.
Therefore, automated methods are increasingly needed to efficient process annotation, ensuring scalability and cost-effectiveness.
Initial automated approaches hire external stronger LLMs to annotate the intermediate process generated by smaller LLMs. Furthermore, Monte Carlo based method reduces the reliance on external stronger LLMs, and can use weaker LLMs to complete data annotation and thereby train stronger LLMs via a self-reinforced manner.

\textbf{Annotation with stronger LLM}:
As a straightforward automated labeling method, Luo \textit{et al.}~\cite{luo2023wizardmath} design to utilize a more powerful external model to annotate the intermediate results of a generative model’s inference process.
Rather than relying on manual annotation, the method employs a pre-trained, high-performance model, like GPT series, to evaluate each generated step.
By leveraging the capabilities of a stronger external model, this approach enhances both the accuracy and scalability of the labeling process, making it more feasible for large-scale tasks.
However, the major limitation of this approach is its reliance on the highly capable external model, which means the performance of the labeling process is ultimately constrained by the capabilities of the external model used.

\textbf{Annotation by Monte Carlo simulation}:
To reduce reliance on the powerful external models, Wang \textit{et al.}~\cite{wang2024math} and Wang \textit{et al.}~\cite{wang2024multi} propose an improved method that avoids directly scoring the intermediate steps. Instead, their approaches use an external model to continue the reasoning for several steps from the given intermediate output and randomly repeat this simulation process multiple times.
The quality of the intermediate step is then assessed based on the average outcome of these extended inferences.
This Monte Carlo method has shown promising results in tasks such as mathematical problem solving and code generation.

\textbf{Annotation by tree search simulation}:
The approach of using multiple-step Monte Carlo simulation with an external model to assess the quality of intermediate steps based on the average outcomes has become one of the most widely used methods for automated process annotation.
To further enhance the efficiency of this method, Luo \textit{et al.}~\cite{luo2024improve} proposes an improvement by replacing the repeated Monte Carlo simulations with a Monte Carlo Tree Search (MCTS) strategy.
In this improved method, multiple leaf nodes representing the final inference results are generated from the intermediate step using MCTS.
The quality of the intermediate step is then evaluated based on the average outcomes of these leaf nodes.
Compared to random repeated inferences, MCTS leverages tree search to improve the inference quality, while also allowing leaf nodes to share high-quality parent nodes, reducing computational overhead and increasing efficiency.
This method has demonstrated superior performance in mathematical problem solving, outperforming human annotations.

One step forward from the MCTS-based simulation, Zhang \textit{et al.}~\cite{zhang2024rest} introduces a self-refining mechanism into the process annotation.
They leverage the obtained process annotations to train a process reward function (PRM), which in turn improves the performance of the large language model (LLM).
The refined LLM is then used to repeat the MCTS-based simulation, generating higher-quality annotations.
This iterative process, involving repeated cycles of improvement, results in progressively enhanced process annotations.
This method has shown excellent performance across several tasks, including mathematical mathematical problem solving, questioning and answering, and multi-domain knowledge reasoning, demonstrating its effectiveness in continuously refining and improving the quality of the annotations through iterative enhancement.
\section{Learning to Reason: from Supervised to Reinforcement Fine-tuning}
\label{sec::train}

While pre-trained models excel across various tasks, they often struggle with complex reasoning and aligning outputs with human expectations. Fine-tuning is crucial to address these limitations, refining a model’s performance on specific tasks and enhancing its reasoning capabilities. Initially, supervised fine-tuning (SFT) is used, where models learn task-specific patterns from labeled datasets. However, as reasoning challenges grow, methods like reinforcement learning (RL) and Direct Preference Optimization (DPO) offer a more effective approach, using reward models to more efficiently align the model’s output with human-like reasoning, fostering more coherent, responsible, and contextually aware outputs.

\subsection{Optimizing Pre-trained LLM: Supervised Fine-tuning}

Supervised Fine-Tuning is a learning technique that refines pre-trained models' capabilities for specific tasks or domains using labeled data, while retains model's understanding on pre-trained knowledge. While pre-training allows models to learn broad, general-purpose features from massive amounts of unstructured data, fine-tuning specializes the model by exposing it to smaller, task-specific datasets with clear input-output mappings.

SFT is a critical step in improving the reasoning ability of LLMs, enabling their application in downstream tasks by adapting them from general-purpose systems to domain-specific tools. For example, LLMs like GPT~\cite{radford2018improving}, BERT~\cite{devlin2018bert}, and T5~\cite{raffel2020exploring} are pre-trained on vast amounts of text data using self-supervised learning, equipping them with broad language understanding and generation capabilities. However, their outputs are not always aligned with task-specific requirements. Without fine-tuning, LLMs tend to perform poorly on certain reasoning tasks, such as object counting~\cite{zhang2024good}, satellite understanding~\cite{mall2023remote}, and engineering questions answering~\cite{wang2024mmlu}. Through SFT, we can partially address these challenges by refining the model's outputs based on labeled task-specific datasets.

However, the direct application of SFT may not fully explore the model's reasoning capabilities in the desired domains, particularly in tasks that require more complex decision-making or multi-step problem-solving. The introduction of CoT techniques~\cite{wei2022chain} has revolutionized the SFT process, by explicitly training the model to generate intermediate reasoning steps before arriving at an answer. With CoT-based SFT, LLMs are encouraged to generate intermediate reasoning steps explicitly, thus enhancing their reasoning ability to tackle tasks that require more structured and organized thoughts. For instance, ReasonBert~\cite{deng2021reasonbert} shows that fine-tuning models with reasoning chains significantly enhances their performance on tasks such as math word problems and logical reasoning by incorporating step-by-step reasoning processes. Another key study~\cite{lobo2024impact} investigates how fine-tuning models with reasoning improves their interpretability and reduces errors in complex decision-making scenarios by generating more transparent, stepwise thought processes. By fine-tuning with CoT, models not only improve their final answers but also enhance their ability to ``think through'' the problem, providing clearer insights into the model's reasoning process. 

Despite the diverse methods and outstanding performance of SFT, it comes with several limitations. First, SFT heavily relies on high-quality labeled datasets, which can be expensive and time-consuming to curate, especially for niche domains or tasks requiring expert annotations. Second, SFT may lead to catastrophic forgetting, where the model loses some of its pre-trained general-purpose knowledge during the fine-tuning process, reducing its utility for tasks reasoning outside the fine-tuning domain. Finally, the computational cost of fine-tuning large-scale models can still be prohibitive, even with parameter-efficient approaches, posing challenges for organizations with limited resources. Addressing these limitations requires careful dataset curation, regularization techniques, and the exploration of alternative methods, such as prompt tuning or multi-task fine-tuning, to balance task specialization and generalization.


\subsection{Optimizing Pre-trained LLM: Reinforcement Learning}  

\begin{figure}
    \centering
    \includegraphics[width=0.9\linewidth]{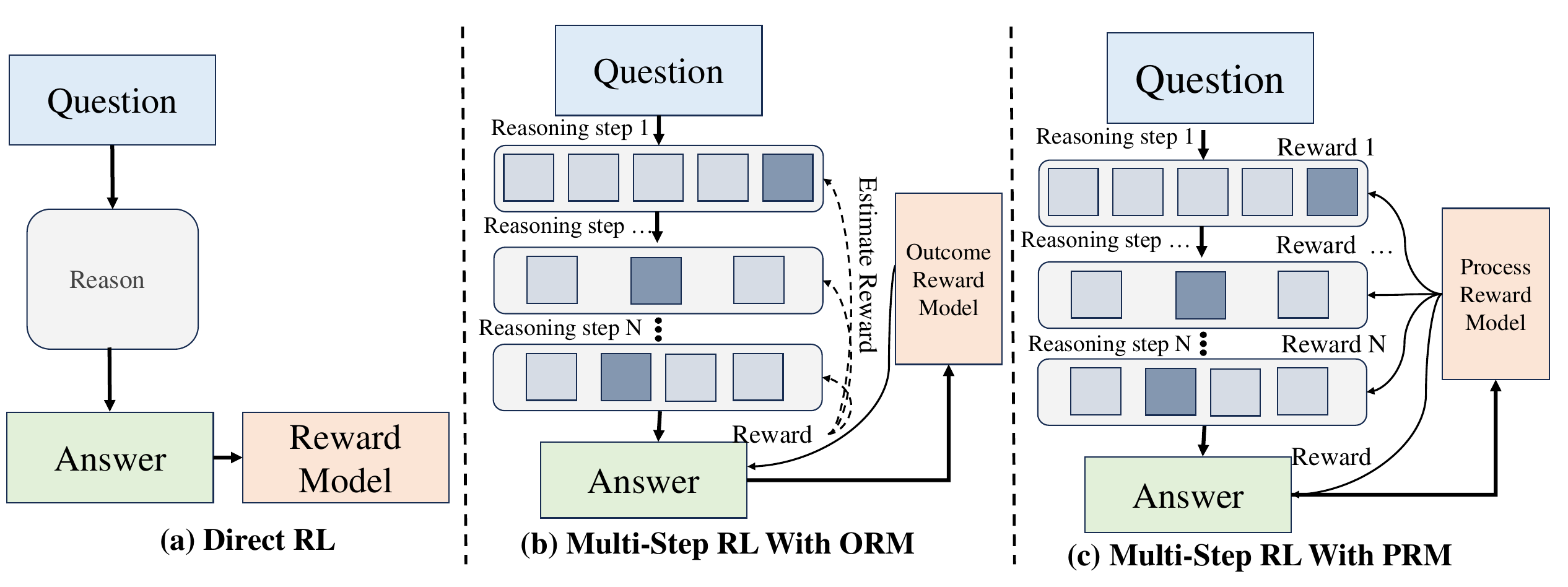}
    \caption{Reward models for Train-time Reinforcement of LLM Reasoning.}
    \label{fig:post-training}
\end{figure}

Due to the high reliance on expensive, high-quality labeled datasets, and high computational costs of SFT, reinforcement learning has emerged as a powerful alternative framework for training models to master reasoning processes. Unlike supervised learning, RL enables models to learn through trial and error reward signals, discovering optimal strategies for achieving specific objectives. As shown in Figure~\ref{fig:post-training} (a), the model takes action based on its current state and receives feedback in the form of a reward signal. This feedback guides the model to update its parameters over time, optimizing for cumulative rewards. 

\textbf{Classic reinforcement learning.}
RL has become a critical step in the development of LLMs. In RL framework, the parameters of LLMs are updated based on the rewards for their actions. Specifically, the value function or Q-function is updated based on the feedback of reward model, attributing the credit for an action’s outcome entirely to its immediate effect. This approach simplifies the framework, making it conceptually straightforward while enhancing the model's ability to respond effectively. Two key methods currently dominate RL training for LLMs: Reinforcement Learning from Human Feedback (RLHF) and Reinforcement Learning from AI Feedback (RLAIF).

Ouyang \textit{et al.}~\cite{ouyang2022training} use RLHF to align LLMs with human intent. Besides, by fine-tuning GPT-3 on human-labeled demonstrations and rank comparisons, they develop a reward model predicting human annotator's preferences. It effectively aligns trained LLMs with human preferences, outperforming GPT-3 in reasoning and instruction-following despite being smaller. Bai \textit{et al.}~\cite{bai2022training} also leverage RLHF to create helpful and harmless language models. Following a Helpful, Honest, and Harmless framework, they fine-tune a base model, train a preference model with rejection sampling, and iteratively refine it with human feedback. This process produces AI assistants that excel in NLP tasks and demonstrate strong ethical reasoning.

To reduce the reliance on large human-labeled datasets, Bai \textit{et al.}~\cite{bai2022constitutional} propose Constitutional AI, a framework for training AI assistants to be helpful and harmless using principles instead of expensive human feedback. The process includes two phases: supervised learning and RLAIF. In the supervised phase, the model critiques and refines its outputs based on constitutional principles, creating a fine-tuning dataset. In the RLAIF phase, the model generates self-assessments to guide training, bypassing the need for human-labeled data on harmfulness. 
Ramamurthy \textit{et al.}~\cite{ramamurthy2022reinforcement} focus on using RL to align LLMs with human preferences. They introduce RL4LMs, a library for RL-based fine-tuning, and the GRUE benchmark, which evaluates models using reward functions reflecting human preferences. To address training challenges, they propose the natural language policy optimization algorithm, which stabilizes training by constraining token sampling. This work provides a strong foundation for integrating RL into LLM fine-tuning for improved alignment and performance.

\textbf{Direct preference optimization}
Classic RL methods rely on training a reward model to score outputs based on human preferences. While DPO streamlines this process by directly leveraging preference data without requiring an explicit reward model. Instead of optimizing a complex reward function, DPO uses pairwise preference comparisons, \emph{i.e.,} data indicating which of two outputs is preferred by humans. This direct approach simplifies the learning pipeline while preserving the alignment benefits of RL-based methods, which is often simpler and more effective. Rafailov \textit{et al.}~\cite{rafailov2024direct} introduce DPO, a novel framework for aligning language models, which directly optimizes the policy to align with human preferences through a simple classification loss. By parameterizing the reward model to derive an optimal policy in closed form, DPO eliminates the need for sampling and extensive hyperparameter tuning during fine-tuning. Experiments show that DPO matches or surpasses RLHF methods like PPO in tasks such as sentiment control, summarization, and dialogue generation, while being more stable, computationally efficient, and effective in producing reasoning outputs. Amini \textit{et al.}~\cite{amini2024direct} propose Direct Preference Optimization with an Offset (ODPO), an extension of DPO for aligning language models with human preferences. ODPO improves upon DPO by considering the degree of preference between responses rather than treating all preference pairs equally. It introduces an offset in the likelihood difference between preferred and dispreferred responses, proportional to their quality difference. This approach not only improves alignment but also strengthens the model's reasoning capability, particularly in tasks such as sentiment control, toxicity reduction, and summarization. Experiments demonstrate that ODPO achieves better alignment and responsible behavior, especially when preference data is limited.

In conclusion, the RL and DPO methods offer a straightforward and effective method for fostering reasoning ability in LLMs. By focusing on immediate rewards following each action, these methods also align models with human preferences. The emphasis on short-term feedback simplifies the learning process, avoiding the complexities of credit assignment across long sequences. This streamlined approach is particularly well-suited for real-time applications and tasks requiring clear, concise reasoning, ultimately strengthening the ability of LLMs to deliver coherent and ethical outcomes.

\subsection{Enhancing Multi-step Reasoning with Outcome Reward Model}  


For complex reasoning tasks, such as mathematical problem-solving, LLMs need to perform multi-step reasoning like Chain-of-Thought to ultimately reach an accurate solution. In these tasks, the reward feedback is typically only available after all the reasoning steps are completed and the final solution is obtained. As shown in Figure~\ref{fig:post-training} (b), this is known as the outcome reward model (ORM). In such cases, the key to improving the LLM’s reasoning ability lies in distinguishing the correctness and importance of intermediate reasoning steps based on the outcome rewards.


\textbf{Classic reinforcement learning.} 
ReFT~\cite{trung2024reft} applies the PPO~\cite{schulman2017proximal} method from RLHF~\cite{ouyang2022training} to reasoning tasks. Based on the outcome reward model, the value function in PPO is able to infer the contribution of intermediate reasoning steps. Compared to supervised fine-tuning, ReFT is capable of learning more diverse reasoning paths, exhibiting stronger generalization abilities in reasoning tasks. However, VinePPO~\cite{kazemnejad2024vineppo} discovers that the value network trained with ORM in PPO exhibits significant bias when identifying the value of intermediate reasoning steps, a well-known challenge in RL called the credit assignment problem. To address this issue, VinePPO abandons the value network in PPO and instead employs a Monte Carlo sampling method to compute unbiased estimates of the value function. Experimental results demonstrate that VinePPO consistently outperforms typical PPO in mathematical reasoning tasks. Critical Plan Step Learning (CPL) is a method designed to enhance LLM's generalization in reasoning tasks by searching within high-level abstract plans~\cite{wang2024cpl}. CPL employs Monte Carlo Tree Search (MCTS) to explore different planning steps in multi-step reasoning tasks, and utilizes Step-APO to learn critical plan steps. This approach enables models to learn more diverse reasoning paths, thereby improving generalization across various tasks. Subsequently, the model iteratively trains the policy and value models to further enhance performance. During each iteration, the policy model generates plan steps and final solutions, while the value model evaluates the quality of intermediate steps. Training data, generated by MCTS, is used to update both the policy and value models.



\textbf{Direct preference optimization.} 
In the task of mathematical reasoning, directly employing the DPO~\cite{rafailov2024direct} method for preference optimization yields suboptimal results due to the presence of lengthy reasoning steps in the preference data. Amini et al.~\cite{amini2024direct} introduced ODPO, which refines DPO by taking into account the degree of preference between responses instead of treating all preference pairs as equal. ODPO has achieved significant improvements over DPO in mathematical reasoning tasks.

In summary, the primary challenge of training based on outcome rewards lies in distinguishing the correctness and importance of intermediate reasoning steps. Current methods, primarily based on Monte Carlo sampling or Monte Carlo Tree Search, offer advantages in estimating the significance of these intermediate steps, though the computational cost during search remains high. Existing work has primarily focused on mathematical or other reasoning problems, where the final solutions can be easily verified. These methods can be extended to a wider range of reasoning tasks, including those where the solutions are difficult to validate. A potential approach is to learn a reward model based on human annotation data, and use it to judge the quality of the final solution. Based on the final score provided by the reward model, Monte Carlo sampling or search techniques can then be employed to further improve performance.

\subsection{Enhancing Multi-step Reasoning with Process Reward Model}

Process Reward Model (PRM) based Reinforcement Learning represents a significant advancement in LLM reasoning, emphasizing the evaluation of intermediate steps rather than solely focusing on end-state outcomes. As shown in Figure~\ref{fig:post-training} (c), the reward of PRM is distributed across each reasoning step, rather than being concentrated at the final outcomes. By providing nuanced feedback throughout the reasoning trajectory, PRM enables models to optimize behavior with greater alignment to human preferences and complex task requirements. This approach is crucial for tasks that involve sequential decision-making, where intermediate steps or decisions are of significance for the final goal. We explore PRMs' evolution and highlight their role in improving reasoning by providing step-level rewards during complex tasks.

\textbf{Classic reinforcement learning}
A series of recent works apply PRMs for mathematical or logistic reasoning, since a seminal work from OpenAI~\cite{lightman2023let} has proven the importance of process reward. SELF-EXPLORE~\cite{hwang2024self} uses PRMs to enhance mathematical reasoning by identifying and addressing ``first pits'', which are the initial incorrect steps in problem-solving. By rewarding steps that correct such errors, PRMs enable self-supervised fine-tuning without requiring extensive human annotations. This model achieves significant improvements in accuracy on mathematical benchmarks like GSM8K and MATH by leveraging step-level fine-grained feedback. MATH-SHEPHERD~\cite{wang2023math} introduces a PRM framework designed for step-by-step verification and reinforcement in mathematical reasoning tasks. By automating process supervision through MCTS-inspired methods, MATH-SHEPHERD eliminates the need for human annotations while ensuring high accuracy in multi-step problem-solving. PRMs are employed to reinforce logical progression and correctness, resulting in improved performance on benchmarks like GSM8K and MATH. DeepSeekMath integrates PRMs via Group Relative Policy Optimization (GRPO)~\cite{shao2024deepseekmath}, a RL algorithm that optimizes step-level rewards. PRMs are used to enhance mathematical reasoning and reasoning consistency across domains. By focusing on intermediate reasoning steps, DeepSeekMath achieves state-of-the-art performance on several benchmarks, showcasing the power of PRMs in mathematical domains. Scaling Automated Process Verifiers introduces Process Advantage Verifiers (PAVs), a PRM variant, to evaluate step-level progress in problem-solving~\cite{setlur2024rewarding}. PAVs use step-level supervision to improve the efficiency and accuracy of search algorithms and reinforcement learning. By focusing on steps that make meaningful progress toward a correct solution, PAVs enable substantial gains in sample efficiency, compute efficiency, and reasoning accuracy compared to outcome reward models. This demonstrates the importance of fine-grained process rewards in scaling LLM reasoning capabilities.

\textbf{Interactive process reward models.}
PRMs are also applied to interactive tasks, such as conversation and multi-turn question answering. ArCHer employs a hierarchical RL approach using PRMs to train agents for multi-turn, long-horizon tasks~\cite{zhou2024archer}. It implements a dual-layer system: a high-level value function evaluates utterance-level rewards, while a low-level PRM optimizes token-by-token generation within each turn. This hierarchical structure ensures more effective credit assignment and allows for nuanced training of language models to handle multi-turn interactions and reasoning tasks. The use of PRMs enables ArCHer to scale efficiently, achieving significant gains in sample efficiency and performance across agent tasks. Multi-turn Reinforcement Learning from Preference Human Feedback~\cite{shani2024multi} integrates PRMs into multi-turn reinforcement learning to optimize long-term objectives with human feedback. The Multi-turn Preference Optimization (MTPO) algorithm compares entire multi-turn interactions to generate preference signals, where PRMs are used to assign step-by-step rewards. This enables LLM agents to align behavior with long-term goals, improving overall performance in dynamic, multi-turn tasks such as conversations and strategic decision-making. 

\textbf{Direct preference optimization.}
Several recent studies leverage MCTS to enable the optimization of multi-step reasoning tasks through Direct Preference Optimization~\cite{xie2024monte,chen2024step,zhang2024rest,chen2024alphamath}. For instance, SVPO~\cite{chen2024step} employs MCTS to automatically annotate step-level preferences for multi-step reasoning tasks. From the perspective of learning to rank, it trains an explicit value model to replicate the behavior of an implicit reward model. Furthermore, SVPO integrates the explicit value model with DPO, where the value model not only aids the policy model in navigating more efficient reasoning paths but also guides preference learning. However, these works primarily focus on first collecting preference data or training a reward model and then performing policy optimization based on static data and the pre-trained reward model. Xie et al.~\cite{xie2024monte} advanced these approaches by integrating data collection and policy preference optimization into an iterative process. This method can be considered as an online version of Direct Preference Optimization, where the updated policy is iteratively utilized to collect preferences through MCTS.

The evolution of multi-step RL techniques for LLMs reflects a transition from sparse outcome-based feedback to detailed process-oriented supervision. PRMs now stand as the centerpiece of the progression of LLM's reasoning ability, offering nuanced, step-level rewards that drive substantial improvements in reasoning tasks. Future research may focus on refining these models and expanding their applicability across diverse task domains.

\subsection{Reinforcement Fine-tuning}


Reinforcement fine-tuning (RFT)~\cite{openai-RFT} is a technique recently proposed by OpenAI for customizing expert LLMs tailored to specific vertical domains. Currently, RFT remains part of a research program and the technical details have not been fully released. Available information suggests RFT utilizes a small amount of preference data provided by users along with a grader model to evaluate LLM's output. This technique enables iterative optimization of the LLM's multi-steps reasoning capabilities. As a result, RFT technique can enhance LLM's strategy of reasoning through similar problems in the optimized domains.

\textbf{The grader model.} 
RFT introduces the concept of a grader model to assess the outputs of LLMs. Considering that reinforcement learning training typically requires a reward model to provide feedback, the grader is likely analogous to a reward model, transforming textual inputs (\emph{e.g.}, questions and answers) into scalar values of reasoning quality. This suggests that the grader could act as a reward model trained on user-provided preference data, potentially operating as either an outcome reward model or a process reward model~\cite{lightman2023letsverifystepstep}.

\textbf{Data efficiency.} 
In OpenAI's live sessions, it was mentioned that RFT can enable learning in new domains with as few as dozens of user preference data. This suggests that RFT facilitates the exploration of diverse reasoning paths to address tasks based on limited preference data. Such an approach demonstrates remarkably high sample efficiency while mitigating the risk of overfitting~\cite{blob-RFT}.

\textbf{Training stability.}
The stability of reinforcement learning training is notoriously difficult problem that presents significant challenges to its broader application. Variations in random seeds or adjustments to certain hyperparameters can greatly impact the training outcomes of RL. In the context of the RFT project, OpenAI has announced plans to make this technology available to the public via APIs, enabling users to fine-tune domain-specific expert models using their own data. This claim potentially indicates that RFT has achieved a level of stability sufficient for reliably fine-tuning language models using RL techniques.


\section{Test-time Scaling: from CoTs to PRM Guided Search}
\label{sec::test}

\begin{figure}
    \centering
    \includegraphics[width=0.99\linewidth]{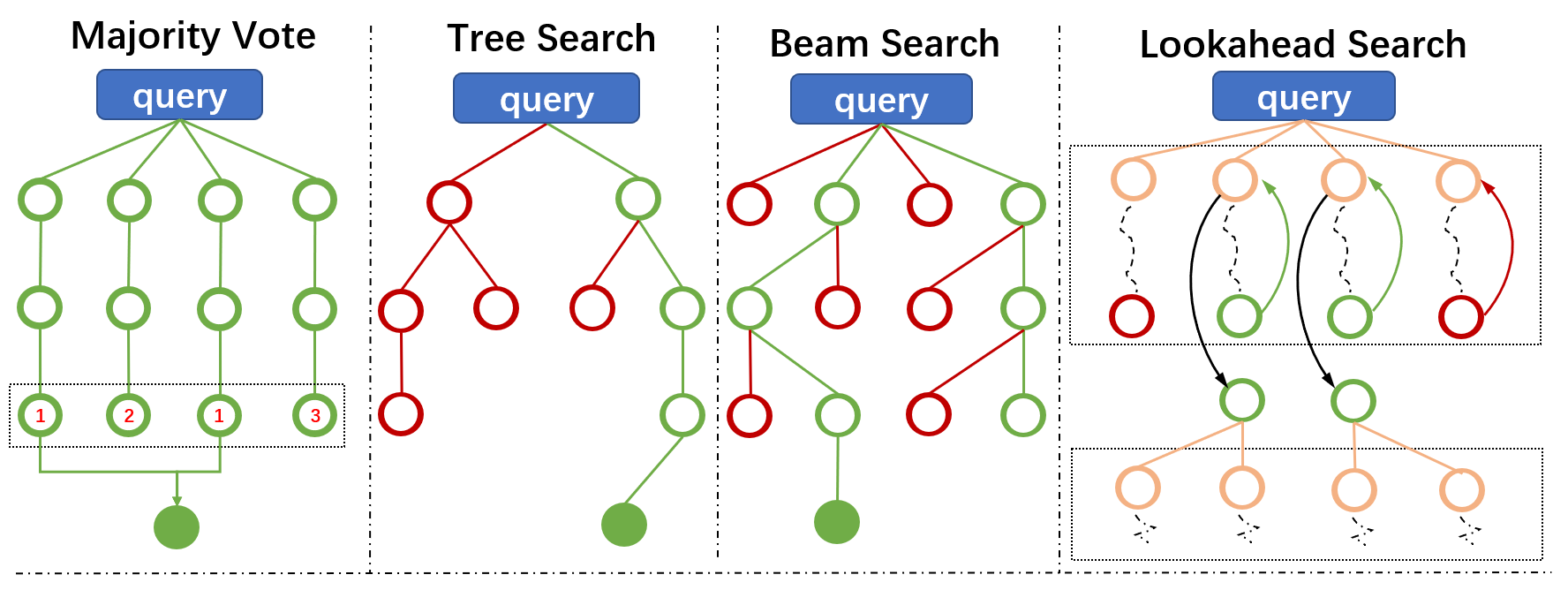}
    \caption{Diagrams of Different Search Algorithms for Test-time Reasoning Enhancement.}
    \label{fig:search}
\end{figure}

\subsection{Elicit Deliberate Thinking with Prompts}
Beyond train-time optimization through techniques such as reinforcement learning, researchers have discovered that test-time prompting techniques like Chain-of-Thought and Tree-of-Thoughts can further enhance LLMs' capabilities~\cite{wei2022chain,wang2022self}. While simply asking models for direct answers often yields suboptimal results, guiding them through explicit reasoning processes at test-time significantly improves their performance~\cite{kojima2022large}. These prompting strategies have shown remarkable effectiveness across various domains, from mathematical reasoning to complex decision-making tasks~\cite{yao2022react,zhou2022least}. The emergence of structured prompting methods like ReAct and Least-to-Most Prompting has demonstrated that LLMs can benefit from explicit guidance in organizing their thought processes, leading to more reliable and interpretable outputs~\cite{zhang2022automatic}. Although these approaches typically increase token consumption and computational overhead, they provide a compelling complement to train-time methods by enhancing LLMs' reasoning capabilities and solution accuracy without requiring model parameter modifications~\cite{yao2024tree,besta2024graph}. This suggests a promising direction for improving LLM performance through sophisticated test-time interventions rather than solely relying on model architecture or training modifications.
\subsection{PRM Guided Search}
As previously mentioned, PRM marks a significant shift from sparse outcome-based feedback to detailed process-oriented supervision. Moreover importantly, PRM can also be utilized during the test-time phase, where it can further boosts the model's reasoning capabilities. OpenAI o1 series models stand as a prominent example of the advanced application of PRM. 
The new test-time scaling laws suggest that inference capabilities can be effectively enhanced by increasing test-time compute, providing a clear direction for the future development of LLMs. We introduce some methods applied during the inference phase, as shown in Figure~\ref{fig:search}. Red hollow circles represent the reasoning paths discarded during the algorithm's exploration process in the inference phase, green hollow circles signify the reasoning paths adopted during exploration, and green solid circles mark the endpoints of the reasoning paths once the correct answer is identified.

\textbf{Majority Vote:}
Majority vote is one of the most straight-forward strategy to generate one final answer from dense test-time compute. During inference, each inference trace produces a prediction for a given input. The basic idea is to select the answer which most inference traces accord with. The predictions from all the models are then aggregated, and the class that appears the most (the "majority vote") is selected as the final output: $f* = argmax_f \sum_y \mathbb{I}_{final\_ans(y)=f}$, where $\mathbb{I}$ is the indicator function and $y$ is each evaluation trace. 


\textbf{Tree Search~\cite{browne2012survey}:}
Tree Search is a classic algorithm that systematically explores different choices by recursively constructing a search tree. It is commonly used in complex decision-making problems, such as board games and planning tasks.
Monte Carlo Tree Search (MCTS) is one of the most widely used tree search methods. It consists of four main steps: Selection, Expansion, Simulation, and Backpropagation. By progressively expanding the search space, MCTS incrementally improves decision-making.
Tree Search has already been applied in some LLM inference tasks, achieving notable success. For instance, the Tree-of-Thoughts framework~\cite{yao2024tree} enables LLMs to consider multiple reasoning paths structured as a tree. It incorporates self-evaluation to make thoughtful decisions, determining the optimal course of action for the next step. This approach significantly enhances the performance of model inference.

\textbf{Beam Search~\cite{smoke1961program}:}
Beam Search is an improved version of greedy search, commonly used in generation tasks to select the optimal output sequence. The main idea is to retain the top-K highest-scoring paths (referred to as beams) at each time step from all candidate paths for further expansion. Unlike greedy search, Beam Search maintains multiple candidate paths, thereby expanding the search space and improving generation quality.
Beam Search is widely applied in LLM inference. For example, BART~\cite{lewis2019bart} uses Beam Search as its primary inference strategy, demonstrating its outstanding effectiveness in text generation tasks.

\textbf{Lookahead Search~\cite{snell2024scaling}:}
Lookahead Search is another promising method that has the potential to significantly enhance LLM inference. It modifies the scoring mechanism at each step of Beam Search. Instead of selecting the best candidates based solely on the scores of the current step, Lookahead Search performs forward simulations by rolling out up to \( k \) steps ahead. If a solution endpoint is reached during the forward simulation, the process halts early. During Lookahead Search, a pre-trained and frozen Predictive Reward Model is used to score each step of the simulation. The cumulative scores derived from the PRM over the \( k \)-step simulation are then used to determine whether to retain or discard a beam branch. This strategy improves decision-making by incorporating more context into each evaluation step. Compared to beam search, lookahead search increases the depth of the exploration space, allowing for the judgment of current decision-making based on the results of more distant simulated decisions. However, it also increases the demand for computational resources which also leads to poor performance when computational resources are limited.

\section{Path toward Large Reasoning Model}
\label{sec::implementation}


\subsection{Development of OpenAI o1 Series}

In September 2024, OpenAI released o1, a groundbreaking language model that represents a significant advancement in AI reasoning capabilities, particularly excelling in complex tasks like mathematics, coding, and scientific problem-solving. On December 20, 2024, OpenAI opened testing applications for o3, an upgraded version of o1~\cite{openai2024o3}, which is considered to have literal doctorate-level intelligence~\cite{openai2024o32}. These model achieve remarkable results across various challenging benchmarks, including scoring at the gold medal level in International Mathematics Olympiad~\cite{li2024openai} and matching PhD-level performance in physics, chemistry, and biology questions~\cite{hayawi2024cross}. Extensive evaluations show distinct reasoning patterns of o1 series through systematic analysis of its basic reasoning capabilities. We list the key findings of existing research as follows:

\paragraph{Effective knowledge integration.} Initial comprehensive evaluations~\cite{zhong2024evaluation} demonstrate o1's structured analytical approach and knowledge integration in fundamental problem solving tasks, achieving 83.3\% success rate in competitive programming through step-by-step logical deduction, where the model demonstrates clear ability to use their knowledge to decompose complex problems and follow formal derivation processes. The model's structured understanding and interconnected knowledge application is further evidenced in specialized fields like radiology and chip design, where accurate diagnosis and complex circuit analysis require integration of multiple domain concepts. Systematic assessments~\cite{latif2024systematic} quantitatively validate this pattern, showing 150\% of human-level performance in structured analytical thinking and computational reasoning tasks. This advantage is particularly prominent in scenarios requiring knowledge integration across domains, such as applying physical principles to biological systems or combining statistical methods with domain-specific constraints, indicating a fundamental capability in knowledge synthesis and application.

\paragraph{Systematic problem decomposition.} o1 maintains consistent performance across tasks of varying complexity levels, showing systematic problem decomposition in handling increased difficulty. In mathematical reasoning, detailed studies~\cite{de2024system} show its systematic problem decomposition approach, achieving near-perfect scores on the Dutch Mathematics B exam through structured solution steps. The model demonstrates ability to identify key mathematical principles, construct formal proofs, and verify solution validity step by step. This consistency extends to more complex scenarios, as validated by research~\cite{davis2024testing} on 105 science and math problems of increasing difficulty, where the model maintains high accuracy even as problem complexity increases in terms of both conceptual depth and computational requirements. In programming tasks, this pattern is further demonstrated through systematic debugging~\cite{hu2024can} on the QuixBugs benchmark, where o1 maintains consistent performance across bugs of varying complexity through a structured three-step approach: error identification, root cause analysis, and targeted correction.

\paragraph{Reliable and coherent reasoning in complex tasks.} The model's reasoning adapt effectively across different problem types, always showing consistence of reasoning chains in various tasks. In planning tasks, PlanBench evaluations~\cite{valmeekam2024llms} demonstrate its systematic handling of both deterministic and probabilistic scenarios, showing significant improvement in constraint satisfaction and state management. The model shows particular strength in handling problems with incomplete information and dynamic constraints, maintaining consistent performance in both standard and rare task variants~\cite{mccoy2024language}. This adaptability indicates robust generalization capabilities across different problem formulations. Studies on complex planning~\cite{wang2024planning} further show o1's ability to maintain reasoning coherence in long-horizon tasks, effectively managing extended dependency chains and context transitions. This is evidenced by its performance in multi-step planning problems where intermediate goals must be correctly sequenced and dependencies carefully managed, demonstrating advanced capabilities in temporal reasoning and causal understanding.

\paragraph{New Scaling Laws for Large Reasoning Models.} Empirical studies demonstrate o1's distinctive scaling patterns in both training and inference phases. During training, the model's large-scale reinforcement learning algorithm teaches it to think productively using chain of thought in a highly data-efficient process~\cite{openai2024learning}. Research~\cite{snell2024scaling} shows that through optimized test-time computation strategies, the model achieves significant performance improvements across various reasoning tasks. Comprehensive evaluations~\cite{zhong2024evaluation,latif2024systematic} reveal that o1's reasoning capabilities can be effectively enhanced through advanced computation allocation during inference, particularly in complex problem-solving scenarios. The constraints on scaling this approach differ substantially from those of LLM pretraining, with performance consistently improving with more time spent thinking~\cite{openai2024learning}. This is evidenced in programming tasks, where allowing 10,000 submissions per problem enables the model to achieve significantly better results, scoring above the gold medal threshold even without test-time selection strategies. The model's ability to effectively utilize additional computation resources during both training and inference suggests a fundamental advancement in reasoning architecture, demonstrating particular strength in scenarios where traditional approaches might require significantly larger model sizes.



\subsection{Open-source Attempts of Large Reasoning Models}

Open-source frameworks have also made substantial strides in developing advanced reasoning capabilities for LLMs. These frameworks serve as invaluable references for researchers and developers aiming to replicate or approximate the reasoning strengths of proprietary models like OpenAI's o1. In this section, we introduce four significant open-source efforts, each of which employs distinct strategies to enhance LLM reasoning (summarized in Table~\ref{tab:osModel}). By exploring their unique implementations, we aim to provide insights into the diverse methodologies used to reinforce reasoning abilities in LLMs.

\begin{table}[h]
\centering
\small
\caption{Open-source Attempts of Large Reasoning Models: A Contribution Point of View.}
\label{tab:osModel}
\begin{tabular}{|c|c|c|c|c|}
        \hline
                                  & Data Construction & Pre-Training & Post-Training & Test-time Improvement \\ \hline
        OpenR\cite{wang2024openr} & MCTS & - & GRPO & Best-of-N \\ \hline
        Rest-MCTS*\cite{zhang2024rest} & MCTS (in loop) & - & SFT & Tree Search \\ \hline
        Journey Learning\cite{qin2024o1} & - & - & SFT & Tree Search \\ \hline
        LLaMA-Berry\cite{zhang2024llama} & MCTS & -  & DPO & Borda Count \\ \hline
    \end{tabular}
\end{table}


\paragraph{The OpenR project\cite{wang2024openr}.} The project claimed that it is the first open-source framework to explore the core methods of OpenAI’s o1 model with reinforcement learning techniques. The core of OpenR replication is to construct step-by-step reasoning data, where the more precise and fine-grained feedback is obtained instead of purely final answers. The automated data augmentation algorithm OmegaPRM~\cite{luo2024improve} is adopted by selecting reasoning trajectories from a constructed search tree. Based on the augmented process data with supervision on each reasoning step, a process reward model is further trained in a supervised learning scheme, based on a pre-trained Qwen2.5-Math-7B-Instruct model~\cite{yang2024qwen2}. The PRM can be directly deployed during test-time compute, integrated with either majority-vote, best-of-N or beam search methods. It can also be utilized to finetune LLM within the post-training stage using RL. Experiments are conducted to demonstrate the effectiveness of the PRM in test-time compute and post-training each.

\paragraph{Rest-MCTS*\cite{zhang2024rest}.} Rather than training PRM and the finetuned policy model separately, they instead integrate these two updates within one mutual self-training loop. Process reward as supervision for PRM training and reasoning traces for policy model training are collected in advance, based on a similarly designed MCTS algorithm. Then the iterative training process starts based on initial policy $\pi$ and initial PRM values $V_{\theta}$. The policy further iteratively perform the MCTS and generate solutions, while the values influence the tree search process. The updates of them complement each other iteratively. 

\paragraph{The o1 Replication Journey project\cite{qin2024o1}.} Rather than thoroughly considering improvement implementation in both stages, the project aims to replicate OpenAI’s o1 model’s reasoning abilities by focusing on comprehensive training strategies. It emphasizes a structured training diagram that incorporates trial-and-error, reflection, and backtracking to build deep causal reasoning. A core aspect of the project is data generation, with high-quality training examples designed to model complex reasoning paths. Using a journey learning method, o1 Replication Journey exposes the model to varied logical sequences and corrections, encouraging exploration and adaptability within the training stage. However, the o1 Replication Journey is less sophisticated at the inference stage, lacking advanced post-training techniques, which limits its adaptability during real-time reasoning. This focus on training over inference highlights its foundational approach compared to models with dynamic inference optimizations.

\paragraph{LLaMA-Berry\cite{zhang2024llama}.} The project directs its focus on optimizing reasoning abilities at the inference stage, leveraging the LLaMA-3.1-8B architecture to deliver more sophisticated real-time reasoning adjustments. It employs a unique pairwise optimization approach that combines Monte Carlo Tree Search with Self-Refine (SR-MCTS), allowing the model to dynamically explore and refine solution paths during inference. This configuration grants LLaMA-Berry a high level of adaptability, enabling it to tackle complex, open-ended reasoning tasks efficiently and flexibly. A key component of this framework is the Pairwise Preference Reward Model (PPRM), which evaluates solution paths in pairs, ensuring that high-quality reasoning paths are prioritized. LLaMA-Berry’s Enhanced Borda Count (EBC) then consolidates these preference rankings to guide the model’s decision-making, further enhancing its inference-stage sophistication. This robust architecture positions LLaMA-Berry as a leading example of inference-focused reinforcement, distinguishing it from O1 Replication Journey’s training-centric approach.

These four open-source frameworks, not only demonstrate distinct implementation strategies for reinforced reasoning but also play an essential role in improving the understanding of OpenAI’s o1 model. Together, they expand the range of techniques available to the open-source community, advancing the collective goal of developing sophisticated, transparent, and adaptable reasoning models that bring proprietary-level capabilities to publicly accessible systems.





\section{Other Test-time Enhancing Techniques}
\label{sec::inference}
\begin{figure}[htbp]
    \centering
    \includegraphics[width=0.99\linewidth]{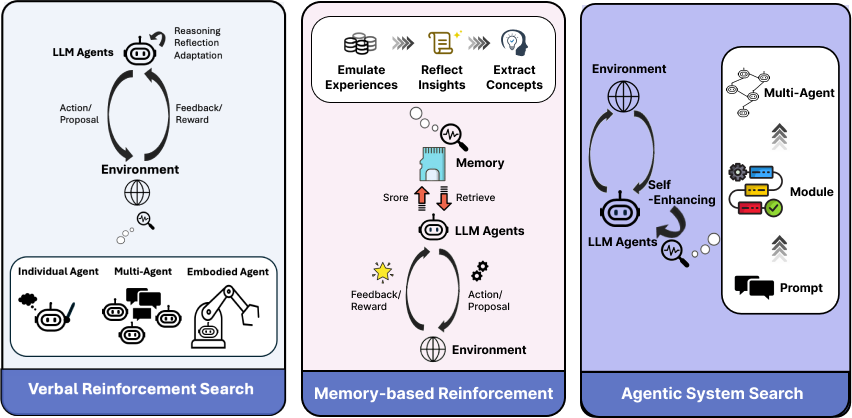}
    \caption{Typical training-free test-time enhancing methods: verbal reinforcement search, memory-based reinforcement, and agentic system search.}
    \label{fig:train_free_test_time_reinforce}
    \end{figure}

In addition to the PRM-guided search, there are numerous other techniques devised to enhance LLM reasoning abilities with more test-time compute. These techniques refine reasoning result dynamically without modifying the model itself. Approaches such as verbal reinforcement search, memory-based reinforcement, and agentic system search, depicted in Figure~\ref{fig:train_free_test_time_reinforce}, demonstrate that substantial reasoning improvements can be achieved with the off-the-shelf LLMs alone. A selection of representative works exploring these methods is summarized in Table~\ref{tab:train-free-test-time-reinforcement-table}. While these methods do not leverage PRM, they offer a foundation for future research to explore hybrid models for further advancing reasoning capabilities.

\begin{table}[h]
\centering
\caption{A list of representative works of training-free test-time reinforcing.}
\label{tab:train-free-test-time-reinforcement-table}
\begin{tabular}{@{}cccccc@{}}
\toprule
      Method             & Category                   & Representative literature                                                                               \\ \midrule
\multirow{6}{*}{Verbal Reinforcement Search} 
                             & Individual Agent   &Romera et al.\cite{romera2024mathematical}, Shojaee et al.\cite{shojaee2024llm},  \\& & Mysocki et al.\cite{wysocki2024llm},Ma et al.\cite{ma2024llm}        &\\ \cmidrule(l){2-4}
                             & Multi-Agent        &Chen et al.\cite{chen2024large},Zhou et al.\cite{zhou2024synergizing},\\ & &Le et al.\cite{le2024multi} ,Yu et al.\cite{yu2024fincon}                 \\ \cmidrule(l){2-4}
                             & Embodied Agent     & Boiko et al.\cite{boiko2023autonomous}
                             \\ \bottomrule
\multirow{6}{*}{Memory-based Reinforcement} 
                             & Experiential Learning  &Zhang et al.\cite{zhang2024large},Gao et al. \cite{gao2024memory} \\& & Qian et al.\cite{qian2023experiential}       &\\ \cmidrule(l){2-4}
                             & Reflective Learning        &Shinn et al.\cite{shinn2024reflexion},Sun et al.\cite{sun2024retrievalaugmentedhierarchicalincontextreinforcement},\\&&Sun et al.\cite{zhao2024expel}                 \\ \cmidrule(l){2-4}
                             & Concept Learning     &Zhang et al.\cite{zhang2024agent},Gao et al.\cite{gao2024self}    ,\\ &&Guan et al.\cite{guan2024richelieu}       \\ \bottomrule
\multirow{4}{*}{Agentic System Search} 
                             & Prompt Level &Madaan et al.\cite{madaan2024self}, Fernando  et al.\cite{fernando2023promptbreeder}\\ & & Yang et al.\cite{yang2024large}        &\\ \cmidrule(l){2-4}
                             &Module Level & Shang et al.\cite{shang2024agentsquare}, Zhang et al.\cite{zhang2024aflow}                 \\ \cmidrule(l){2-4}
                             & Agent Level     &Huot et al.\cite{huot2024agents},Zhuge et al.\cite{zhugegptswarm}       \\ \bottomrule
\end{tabular}
\end{table}
\subsection{Verbal Reinforcement Search}

Verbal Reinforcement Search (VRS) leverages the pre-trained reasoning and semantic capabilities of LLMs to explore and optimize solution spaces. Unlike traditional reinforcement learning or training-intensive approaches, VRS operates purely through test-time inference, using iterative feedback loops to refine solutions without requiring additional training. By drawing on the semantic knowledge encoded in LLMs and their ability to follow complex instructions, VRS provides a versatile approach for navigating diverse problem spaces. This inference-driven framework finds application across individual agents, multi-agent systems, and embodied agents, supporting a wide range of tasks, including programmatic optimization, collaborative decision-making, and interactions in real-world settings. This section analyzes VRS through these three key aspects, delving into the methodologies and unique insights presented within each category.

In \textbf{individual agent settings}, VRS relies on iterative reasoning and feedback mechanisms to refine solutions within structured problem spaces. This approach is well-suited for tasks like mathematical optimization, symbolic reasoning, and hypothesis-driven discovery, where systematic refinement significantly improves problem-solving outcomes. Research on mathematical discovery illustrates how VRS reshapes the problem-solving process into a dynamic iterative cycle. For example, studies on combinatorial problems, including the cap set and online bin-packing, highlight how programmatic solutions evolve through feedback-driven evaluation~\cite{romera2024mathematical}. Similarly, symbolic regression research treats equations as dynamic constructs, iteratively generating, evaluating, and optimizing mathematical expressions~\cite{shojaee2024llm}. These approaches show how VRS navigates constrained spaces, surpassing traditional optimization techniques in efficiency and accuracy. In scientific discovery, VRS has shown its utility in integrating reasoning with empirical data and simulations. Researchers have developed systems for biomedical hypothesis refinement by synthesizing diverse data sources. For instance, applications in oncology use iterative synthesis to address the complexity of multi-scale data~\cite{wysocki2024llm}. In physical sciences, VRS is used to refine hypotheses through simulation feedback, advancing fields like molecular design and the discovery of physical laws~\cite{ma2024llm}. These findings emphasize the role of VRS in connecting abstract reasoning with real-world validation, supporting tasks that are both data-intensive and hypothesis-driven. Reflective processes in heuristic optimization further showcase the flexibility of VRS. For example, researchers have explored the iterative generation and evaluation of strategies for solving combinatorial problems~\cite{ye2024reevo}. This approach focuses on creating adaptive hyper-heuristics that generalize effectively across different domains by continuously refining solutions through feedback cycles. Overall, VRS applies iterative reasoning and feedback to connect abstract problem-solving with real-world applications, addressing challenges in mathematics, science, and optimization with precision and adaptability.

In \textbf{multi-agent systems}, VRS facilitates collaboration between LLM-based agents through natural language communication. These systems leverage shared reasoning and iterative refinement to tackle complex solution spaces, allowing agents to exchange insights and achieve common goals. Meta-structure discovery in heterogeneous information networks (HINs) exemplifies how VRS is applied in multi-agent contexts. Recent research has combined LLM reasoning with evolutionary optimization to refine meta-structures, enhancing their explainability and predictive accuracy~\cite{chen2024large}. Similarly, in socioeconomic prediction, multi-agent systems integrate knowledge graphs and meta-path reasoning to extract cross-task insights for applications like population estimation and economic activity prediction. This approach facilitates collaboration between LLM agents and improves performance in multi-task environments~\cite{zhou2024synergizing}. Causal discovery also benefits from multi-agent frameworks enabled by VRS. For example, systems using LLMs as reasoning agents collaboratively debate and propose causal relationships. By incorporating statistical methods and natural language interactions, these frameworks generate accurate causal graphs while addressing ambiguities in causal relationships~\cite{le2024multi}. In financial decision-making, VRS enhances hierarchical collaboration. The FINCON framework employs a manager-analyst system to refine financial strategies using conceptual verbal reinforcement. By minimizing redundant communication and improving strategy refinement, FINCON demonstrates the utility of VRS in optimizing financial decision-making processes~\cite{yu2024fincon}. With iterative refinement and shared reasoning, VRS supports multi-agent systems in tackling complex tasks such as meta-structure refinement, socioeconomic prediction, and financial decision-making.

In \textbf{embodied agent settings}, VRS is used to address real-world tasks by integrating reasoning with physical interactions, supporting activities such as experimental planning and execution in laboratory settings. These systems extend VRS into dynamic environments, combining semantic reasoning with practical experimentation. For example, autonomous chemical research has demonstrated the use of LLM-powered systems to independently design, execute, and refine experiments~\cite{boiko2023autonomous}. These agents integrate tools such as robotic liquid handlers, spectrometry devices, and web-based research modules to perform tasks like reaction optimization and compound synthesis. One application involves optimizing palladium-catalyzed cross-coupling reactions, where the system uses natural language prompts to determine conditions, calculate stoichiometries, and autonomously execute experiments. When faced with errors, such as incorrect module calls, the system revises its approach by referencing documentation and iterating on the task. This iterative process demonstrates how VRS supports adaptability and precision in experimental workflows. By combining reasoning and real-time feedback, embodied agents illustrate the capability of VRS to refine and optimize complex processes in dynamic environments. These systems reduce human intervention while accelerating scientific discovery, making them a valuable tool for real-world experimentation and innovation.

In general, previous studies have showcased the adaptability and effectiveness of VRS across individual agents, multi-agent systems, and embodied agents. Leveraging the semantic reasoning and iterative feedback capabilities of LLMs, VRS tackles a wide range of tasks without the need for additional training. From structured optimization in mathematical and scientific contexts to collaborative exploration in multi-agent frameworks, and dynamic experimentation in real-world applications, VRS provides a unified approach to problem-solving. VRS as a versatile framework, capable of addressing complex challenges across both computational and physical domains while driving advancements in diverse fields.







\subsection{Memory-based Reinforcement}

When applied to open-ended tasks such as creative writing, complex logical reasoning, and open-world gaming, the solution space tends to expand dramatically, often becoming unbounded or ill-defined. These tasks typically require continuous interaction with the environment to acquire relevant information, making simple solution space searches inefficient. To address these challenges, some studies incorporate an external memory module for LLM agents. This module stores information such as observations, successful and failed actions from past trials. Agents explore their environments iteratively, using memory as a foundation for verbal reinforcement learning. Through this process, they summarize experience, extract interpretable high-level insights of the solution space, and refine their actions in subsequent trials, thereby improving inference performance. These studies not only focus on exploring the external solution space but also emphasize the intrinsic ability of LLM agents to develop an understanding of the solution space from memory. As the agents accumulate memory through environmental exploration, their capabilities are progressively reinforced and generalized to unseen tasks. Specifically, we classify the studies in this area into the following three categories.

\textbf{Experiential learning.} Methods in this category encourage LLM agents to simply emulate favorable experiences stored in memory while avoiding unfavorable ones. REMEMBERER~\cite{zhang2024large} introduces a semi-parametric RL-LLM agent that records past observation-action pairs in memory and uses a traditional off-policy Q-learning algorithm to dynamically maintain and update the Q-value (expected future reward) of each observation-action pair. When faced with a new task, the agent retrieves relevant actions with the highest and lowest Q-values from memory, incorporating these as encouraged and discouraged examples in the prompt. Memory Sharing~\cite{gao2024memory} leverages concepts from multi-agent reinforcement learning to enhance learning efficiency. Multiple agents execute tasks concurrently in a shared environment and contribute high-quality prompt-answer pairs to a collective memory pool. Each agent can retrieve the most relevant examples from this pool to facilitate few-shot learning. Similarly, Experiential Co-Learning~\cite{qian2023experiential} employs a multi-agent framework in which Instructor and Assistant agents alternately provide instructions and solutions during multi-step code generation. This dynamic exchange helps extract shortcuts to reduce redundancy and prevent repetitive mistakes. When encountering new tasks, these agents retrieve relevant memories alternately to improve in-context learning.

\textbf{Reflective learning.} Although using memory as few-shot exemplars is straightforwardly effective, this approach does not fully exploit the semantic comprehension capabilities of LLMs. Some studies argue that LLM agents should reflect directly on the successes and failures stored in memory to summarize underlying causes explicitly, adopting these insights as guidelines. Reflexion~\cite{shinn2024reflexion} is a pioneering effort in this area, reflecting on the reasons behind success or failure semantically based on task feedback signals. It integrates reflective text and past trajectories into prompts to enhance decision-making in subsequent trials. ExpeL~\cite{zhao2024expel} combines imitation and reflection by retrieving the most relevant successful experiences from memory, summarizing patterns of successful trajectories, and identifying insights from comparisons of success-failure pairs. RAHL~\cite{sun2024retrievalaugmentedhierarchicalincontextreinforcement}, inspired by hierarchical reinforcement learning, organizes memory into goal modules and sub-task modules, enabling reflection and experience summarization at different levels. For new tasks, it retrieves relevant experience to formulate high-level goals and low-level sub-tasks separately.

\textbf{Concept learning.} Explicit reflection significantly enhances the inference capabilities of LLMs. Building on this, some studies aim to enable LLM agents to develop generalized ``concepts'' that transcend specific tasks, facilitating a broader understanding of the environment and tasks. This generalization helps agents internalize cognitive abilities from memory and evolve continuously as memory grows. Agent-Pro~\cite{zhang2024agent}, for example, enables agents to establish beliefs about themselves and their environments in card-based gaming. Instead of reflecting on individual actions, it evaluates the rationality and consistency of these beliefs, iteratively refining strategies. Similarly, Richelieu~\cite{guan2024richelieu} equips agents with an understanding of the environment in military strategy games. It retrieves the most relevant states from memory to formulate plans and assess feasibility. By employing self-play, it collects experience autonomously, assuming the roles of all players to advance its knowledge. Self-Evolving GPT~\cite{gao2024self}, inspired by human memory mechanisms, designs a memory-based autonomous learning framework for LLMs. It categorizes tasks to determine relevant memory retrievals and identifies differences between stored memories and the current task to extract shared general experience. Additionally, it generates unseen tasks for practice, consolidating its knowledge based on memory retrieval outcomes.

\subsection{Agentic System Search}
The design of agentic systems plays a crucial role in harnessing the power of LLMs for many downstream tasks. An important branch of test-time enhancing techniques is to leverage LLMs to search the agentic systems. Studies in this area can be classified into three levels of search: prompt level, module level, and agent level. Please note that this approach does not aim to directly search the solution space, rather it leverages the empirical data to optimize the agentic system itself, which is similar to a meta-learning problem. We summarize the related works in this area as follows.

\textbf{Prompt level.} The process of ``verifying and correcting'' improves the prompts by iteratively integrating useful feedback experience. The verification signal can come from external feedback~\cite{gou2023critic}, LLM's self evaluation~\cite{madaan2024self}, and other sources. On the other hand, prompts themselves are also worth searching and optimizing. Automated prompt engineering, such as evolutionary prompt optimization~\cite{fernando2023promptbreeder} and meta prompt iterations~\cite{yang2024large}, can achieve better results than manual prompts, but it also introduces more token consumption.

\textbf{Module level.} Agentsquare~\cite{shang2024agentsquare} proposes to use LLM to search the modular design of agentic system, where the modules are essentially blocks of prompts that have specific functions of Planning, Reasoning, Tool use, and Memory.
The basic units of these agentic modules have standard IO interface that allows them to collaborate well with each other. The advantage of module level search is it allows new agents to easily reused the classic agent design, such as CoT and ToT, through recombination of modules. Besides, Aflow\cite{zhang2024aflow} connects the different calling nodes of LLM through edges represented by code. In addition to the search method, it is necessary to evaluate the performance of the searched agents. The function used to evaluate the performance of agents can also be driven by LLMs to improve search efficiency while closely matching their actual performance.

\textbf{Agent level.} 
ADAS proposes to leverage LLMs to search the entire agentic systems defined in python code space~\cite{hu2024automated}. Besides, multi-agent systems make decisions and achieve goals in a shared environment. In multi-agent level search, key aspects include agent creation, environmental perception, action, interaction, and system evolution. The search for multi-agent systems has achieved good results in downstream tasks such as long story creation~\cite{huot2024agents}. The unified search and optimization mechanism for multi-agent systems is currently being explored. GPTSwarm~\cite{zhugegptswarm} enhances the collaborative capability of agents through graph optimization.

Agentic System Search provides agents with the ability to self-improve, enabling them to optimize themselves to enhance their reasoning abilities without the need to make changes to the LLM structure. The above three levels of search have vast search spaces. The common challenge faced by these three search levels is to improve search efficiency, reduce search costs, and ensure automation while ensuring search rationality.

\subsection{Summary}
The test-time enhancing techniques review in this section currently are not incorporated in the implementations of large reasoning models. However, they have huge potential to further boost the reasoning capacities of LLMs through more comprehensive test-time ``thinking'', facilitating LLMs to strategically reason across the solution space, leverage past experiences and dynamically optimize agentic workflows. Therefore, training LLMs to master these test-time techniques represents a promising future research direction, with the potential to elevate LLMs from ``reasoners'' to fully functional ``agents.''
\section{Evaluation Benchmarks}
\label{sec::benchmark}

\begin{figure}
    \centering
    \includegraphics[width=0.93\linewidth]{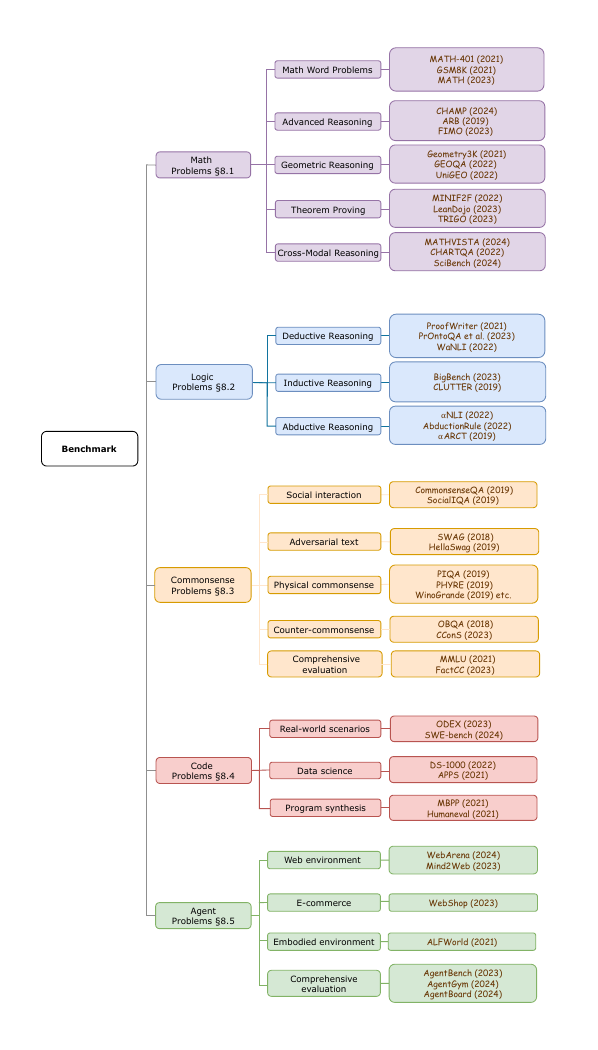}
    \caption{A Taxonomy for LLM Reasoning Benchmarks.}
    \label{fig:taxReason}
\end{figure}

Designing a robust benchmark is important to document the improvement LLM's capabilities. It also plays a crucial role in selecting the promising research direction for further advancement. In this section, we systemically review the popular benchmarks for LLM reasoning, which are summarized with the taxonomy in Figure~\ref{fig:taxReason}. Here, we discuss these benchmarks as follows.

\subsection{Math Problems}
Mathematical reasoning has become a crucial testbed for evaluating LLM's reasoning capabilities. The landscape of mathematical reasoning benchmarks spans from elementary arithmetic to advanced university-level mathematics, providing systematic ways to assess different aspects of mathematical understanding and problem-solving abilities.

In the realm of \textbf{mathematical word problems} (MWP), benchmarks progress from fundamental arithmetic operations to increasingly complex problem-solving scenarios. At the basic level, datasets like MATH-401~\cite{yuan2023largelanguagemodelsperform} evaluate pure arithmetic capabilities through 401 carefully structured expressions, while MultiArith~\cite{roy2016solvinggeneralarithmeticword} and AddSub~\cite{hosseini-etal-2014-learning} assess the ability to translate simple word problems into mathematical operations (such as addition or subtraction). Moving to elementary and high school levels, comprehensive datasets such as GSM8K~\cite{cobbe2021trainingverifierssolvemath} and MATH~\cite{hendrycks2021measuringmathematicalproblemsolving} present more sophisticated multi-step reasoning challenges, with GSM8K offering 8.5K grade school problems and MATH providing 12.5K problems across various mathematical domains with graduated difficulty levels.

The evaluation of \textbf{advanced mathematical capabilities} is primarily conducted through competition and specialized test datasets. Collections like CHAMP~\cite{mao2024champcompetitionleveldatasetfinegrained} and ARB~\cite{amini2019mathqainterpretablemathword} present competition-level problems that require sophisticated problem-solving strategies, while MATHQA~\cite{amini2019mathqainterpretablemathword} incorporates standardized test questions from GRE and GMAT examinations. At the highest level, datasets such as FIMO~\cite{liu2023fimochallengeformaldataset} challenge models with International Mathematical Olympiad problems, testing the limits of automated mathematical reasoning.

\textbf{Geometric reasoning} represents a distinct category requiring spatial understanding and formal mathematical proofs. Datasets like Geometry3K~\cite{lu2021intergpsinterpretablegeometryproblem} and GEOQA~\cite{chen2022geoqageometricquestionanswering} provide specialized geometric problems, while UniGEO~\cite{chen2022unigeounifyinggeometrylogical} offers a unified framework for geometric reasoning tasks focusing on calculation and proving. These benchmarks are particularly valuable in assessing models' abilities to connect visual and mathematical reasoning.

The field of \textbf{theorem proving} and formal mathematics has evolved to include rigorous evaluation frameworks. MINIF2F~\cite{zheng2022minif2fcrosssystembenchmarkformal} and LeanDojo~\cite{yang2023leandojotheoremprovingretrievalaugmented} focus on formal mathematical proofs related to Lean Theorem, while THEOREMQA-MATH~\cite{chen2023theoremqatheoremdrivenquestionanswering} examines understanding of mathematical theorems. Specialized datasets like TRIGO~\cite{xiong_trigo_2023} and PISA~\cite{jiang2021lisa} address specific areas of mathematical reasoning, such as trigonometry and formal proof systems.

Lastly, \textbf{cross-modal mathematical reasoning} has emerged as a crucial area, reflecting the diverse ways mathematical problems are presented in real-world scenarios. MATHVISTA~\cite{lu2024mathvistaevaluatingmathematicalreasoning} and CHARTQA~\cite{masry2022chartqabenchmarkquestionanswering} evaluate visual mathematical reasoning through diagrams and charts, while TABMWP~\cite{lu2023dynamicpromptlearningpolicy} and MultiHiertt~\cite{zhao2022multihierttnumericalreasoningmulti} assess the ability to reason with tabular and textual data. SciBench~\cite{wang2024scibenchevaluatingcollegelevelscientific} bridges the gap between pure mathematics and scientific applications, testing mathematical reasoning in broader scientific contexts.

\subsection{Logical Problems}
Building upon mathematical reasoning capabilities, the ability to engage in systematic logical reasoning stands as another fundamental criterion for evaluating LLM's cognitive abilities. While mathematical reasoning focuses on quantitative operations and formal proofs, logical reasoning encompasses the broader capacity to draw valid conclusions, recognize patterns, and generate rational explanations across diverse contexts. According to Luo et al.~\cite{luo2024logigluebriefsurveybenchmark}, logical reasoning can be classified into three main types: Deductive, Inductive, and Abductive reasoning. Each type represents a distinct cognitive process essential for comprehensive logical analysis while maintaining interconnections in cognitive assessment.

\textbf{Deductive reasoning} also known as premise-based reasoning, involves deriving specific conclusions from general principles with absolute certainty. For example, given a set of rules about relationships between entities, a model must determine what specific relationships must be true. ProofWriter~\cite{tafjord2021proofwritergeneratingimplicationsproofs} exemplifies this category, requiring models to construct explicit logical derivations from given premises. Other benchmarks, such as FOLIO~\cite{han2024folionaturallanguagereasoning} and PrOntoQA~\cite{saparov2023languagemodelsgreedyreasoners} evaluates first-order logic reasoning in natural contexts, WaNLI~\cite{liu2022wanliworkeraicollaboration} have introduced increasingly sophisticated evaluation criteria with 107,885 examples.

\textbf{Inductive reasoning} emphasizes pattern recognition and generalization from specific observations to broader principles~\cite{sep-logic-inductive}. This involves identifying underlying regularities and extending them to new situations, dealing with probability rather than certainty. BigBench~\cite{srivastava2022beyond} with numerous specialized components that examine advanced pattern inference capabilities. Also, CLUTTR~\cite{sinha2019clutrrdiagnosticbenchmarkinductive} benchmark series evaluates this capability through relationship patterns of varying complexity.

\textbf{Abductive reasoning}, also termed explanatory reasoning, refers to the process of forming the most likely explanation for a set of observations or facts, even though the conclusion is not guaranteed to be certain~\cite{sep-abduction}. This type of reasoning tests how models handle scenarios with incomplete information by generating reasonable explanations. The $\alpha$NLI~\cite{nie2020adversarialnlinewbenchmark} benchmark implements this through narrative completion tasks, where models must select the most likely explanation for given situations. The AbductionRule~\cite{young2022abductionrulestrainingtransformersexplain} series offers structured evaluation frameworks across different domains, with specific variants for animal-related and person-related reasoning scenarios. $\alpha$ARCT~\cite{niven2019probingneuralnetworkcomprehension} specifically examines the ability to select and justify plausible explanations and argument comprehension.

\subsection{Commonsense Problems}
Commonsense reasoning remains a significant challenge in NLP, as it aims to evaluate LLM's ability to understand and apply everyday commonsense knowledge. There are various benchmarks targeting different dimensions of commonsense reasoning tasks. For instance, CommonsenseQA~\cite{talmor2019commonsenseqa} requires models to answer reasoning questions grounded in commonsense knowledge bases.

SocialIQA~\cite{sap2019social} focus on \textbf{social interaction} commonsense reasoning, which revolves around causal reasoning in social scenarios. In contrast, datasets like SWAG~\cite{zellers2018swag} and HellaSwag~\cite{zellers2019hellaswag} introduce \textbf{adversarial text} reasoning tasks, where models must predict the most plausible continuation of events based on contextual clues, thereby increasing task complexity. For \textbf{physical commonsense} reasoning, benchmarks such as PIQA~\cite{bisk2020piqa} and PHYRE~\cite{bakhtin2019phyre} concentrate on evaluating models' understanding of everyday physical tasks and interactive reasoning scenarios. PIQA primarily uses question-answering tasks, while PHYRE emphasizes interactive physical simulations. Similarly, WinoGrande~\cite{sakaguchi2020winogrande} builds upon the Winograd Schema Challenge by introducing a larger-scale dataset and mo   re complex disambiguation tasks to test semantic understanding and conference resolution capabilities.

Other works, such as OBQA~\cite{mihaylov2018can} and CConS~\cite{kondo2023probing}, explore model performance in \textbf{counter-commonsense} contexts, highlighting the challenges faced by current models in implicit reasoning and background knowledge utilization. More recently, \textbf{comprehensive benchmarks} like MMLU~\cite{hendrycks2021measuring} and critical studies such as FactCC~\cite{laban2023llms} have further analyzed LLM's commonsense reasoning and factual reasoning. These benchmarks offer valuable perspectives on the generalization abilities of language models and serve as valuable tools for evaluating and improving their performance across diverse commonsense reasoning tasks.

\subsection{Coding Problem}

The development of code generation benchmarks has been instrumental in evaluating the reasoning capabilities of LLMs in programming tasks. These benchmarks assess models' proficiency in generating accurate, efficient, and reliable code across various domains. For example, ODEX~\cite{wang2022execution} introduces an execution-driven evaluation framework for open-domain code generation, emphasizing the importance of running generated code to verify its correctness and functionality. 

As for the \textbf{real-world scenarios}, SWE-bench~\cite{jimenez2023swe} focuses on real GitHub issues, challenging models to resolve practical software engineering problems. In the realm of \textbf{data science}, DS-1000~\cite{lai2023ds} presents a benchmark featuring authentic and dependable data science code generation tasks, enabling the assessment of models' abilities to handle complex data manipulations and analyses. Besides, the APPS benchmark~\cite{hendrycks2021measuring} measures coding challenge competence by evaluating models on a diverse set of programming problems, reflecting the challenges encountered in competitive programming and technical interviews.

MBPP~\cite{austin2021program} focuses on \textbf{program synthesis} problems, assessing models' abilities to generate correct and efficient code based on given specifications, thereby contributing to the understanding of LLMs' capabilities in automated code generation.
The HumanEval~\cite{chen2021evaluating} evaluates LLMs trained on code by providing a set of Python programming problems, each provided with a function definition and accompanying documentation, requiring models to generate correct and functional code solutions.

\subsection{Agent Problems}

The emergence of agent-based benchmarks has revolutionized our ability to assess LLMs as independent agents within interactive environments. These sophisticated evaluation frameworks assess crucial capabilities including decision-making, reasoning, and environmental interaction across diverse scenarios.

WebArena~\cite{zhou2023webarena} provides a practical \textbf{web environment} for building and testing autonomous agents, enabling the evaluation of LLMs' web navigation and interaction skills. Similarly, Mind2Web~\cite{deng2024mind2web} aims to develop generalist agents capable of operating across diverse web tasks, emphasizing adaptability in dynamic online environments.

In \textbf{E-commerce} settings, WebShop~\cite{yao2022webshop} introduces a platform for scalable real-world web interaction, focusing on grounded language agents that can perform tasks such as online shopping, thereby testing models' practical application abilities.
To bridge textual and embodied environments, ALFWorld~\cite{shridhar2020alfworld} aligns text-based inputs with interactive learning scenarios, facilitating the assessment of models' abilities to transfer knowledge between different modalities.

\textbf{Comprehensive evaluation} frameworks like AgentBench~\cite{liu2023agentbench} and AgentGym~\cite{xi2024agentgym} have been developed to systematically assess LLMs functioning as agents. AgentBench includes diverse environments to assess reasoning and decision-making skills, while AgentGym focuses on evolving LLM-based agents across diverse settings, emphasizing adaptability and learning efficiency. Furthermore, AgentBoard~\cite{ma2024agentboard} offers an analytical platform for evaluating multi-turn LLM agents, providing insights into their performance over extended interactions and highlighting areas for improvement in sustained reasoning tasks.  

\section{Discussion}
\label{sec::discussion}

\subsection{Inspirations from the recent advances}

\noindent \textbf{Scaling law of post-training phases.} The inspiration from OpenAI o1 series leads to a new understanding of the pre-training/post-training/inference stages. Particularly, it involves the introduction of self-play reinforcement learning and the process reward learning of high-quality Chain-of-Thought labeled data during the post-training stage. Further, it extends to the scaling law in the post-training stage, which provides inspiration for the difficulties in the further development of the Scaling Law in the training stage. 
As we know, the scaling law in the pre-training and training stages has led to the success of popular LLMs, with the huge investment of training data and computation resources.
However, it now reaches the bottleneck, and thus, the scaling law of post-training phases may be the driving force for the next period of development of large language models.
Furthermore, LLM-driven agents~\cite{xi2023rise} has also shown great potential with carefully-designed workflow even if the reasoning abilities have not been reinforced. Therefore, it is still an open question whether there will also be a similar scaling law regarding resource consumption and performance in LLM agents, which could be the potential to further enhance LLM in real-world applications.
Lastly, there may be a relationship between the currently exhibited test-time scaling law and the model's ability to follow instructions; that is, it must have a sufficiently strong instruction following ability to demonstrate test-time scaling laws. For example, the success of verbal reinforcement search techniques require the LLMs to have the basic ability to follow instructions. Thus, if the LLMs cannot accurately follow instructions, the complicated post-training techniques might not work properly.

\noindent \textbf{Generating high-quality data through searching.} 
Both the technical ideas of OpenAI o1 series disclosed by its core technical personnel and the open-source works attempting to reproduce OpenAI o1 currently regard the generation of high-quality data (including CoT data) as the key point, although different approaches such as Monte Carlo Tree Search, LLM generation, and others have been adopted.
That is, the development of large reasoning models reaches a stage where high-quality process reward data is more important than general pre-training data size. Similarly, as discussed above, it may inspire us to refer to these related approaches in LLM agents as well, first to conduct high-quality data generation and then enhance the learning of slow reasoning as well as the acquisition of capabilities.

\subsection{Slow-thinking and reasoning}

Even if the breakthrough of OpenAI o1 series at the engineering level remains unknown, theoretically and technically, its breakthrough currently seems to mainly lie in the post-training learning of slow-thinking data.
Also, the human cognitive science of ``System 1 + System 2'' has been repeatedly mentioned, but the ideas on how to implement it based on large models have been constantly updated, mainly still staying at the stage of drawing on the concept of slow thinking. 
That is, the mechanism named ``System 1 + System 2'' of human brains has guided the design of LLMs, but the guidance is still very limited. In other words, the imitation of the human brain is only about system-level design rather than very detailed techniques. The complex mechanisms of human slow thinking and their benefits still show high potential to support the next-level reasoning abilities of LLMs. To accomplish it, the domain knowledge of slow thinking should be used in the related designs such as reasoning data generation, reward functions, learning process, etc.

There has been no truly significant and representative work on the theoretical analysis of slow-thinking of LLMs up to now. The generative artificial intelligence is so mysterious that understanding LLMs also requires some tricks or special techniques such as new metrics for understanding hallucination of LLM~\cite{farquhar2024detecting}. 
To understand the slow-reasoning abilities, we may need to also step into the theoretical analysis.
Taking the two different versions, OpenAI o1 Preview and OpenAI o1 Mini, as examples, the main difference lies in the cost and depth of thinking in the CoT inference stage, yet they show significant differences in tasks such as text generation, code generation, and mathematical problem-solving. 
The special characteristics of reasoning shown by LLMs also inspire us to design task-adaptive usage and applications. Specifically, it may support more interesting insights to link the reasoning mechanism and the performance in different tasks.

\subsection{Downstream applications and open problems}

As pointed out throughout this paper, the progress of reasoning enhancement technologies is rapid. 
The reasoning abilities are not limited to the tasks in these popular benchmark tasks, but also in more general tasks in downstream applications. For example, the FunSearch work~\cite{romera2024mathematical} has shown the general ability for
tasks difficult to provide solutions, but verification is fast. 
There could be a bundle of tasks with similar features in various domains, such as urban planning, logistics scheduling, etc.
An interesting question is whether there may be many complementary problems in current research, which is difficult to verify, but the reasoning process is easier. It may be possible to further verify the quality of some answers by combining LLMs and external evaluators, or we can use these answers with evaluated scores to train the reward model.

\section{Conclusion}\label{sec::conclusion}


The recent evolution of LLMs has significantly advanced their human-like reasoning capabilities. 
The innovations of introducing concepts like ``thought'' as intermediate steps, leveraging reinforcement learning techniques for train-time scaling, and using search algorithms for test-time scaling have laid the groundwork for large reasoning models, which can address increasingly complex cognitive tasks, as exemplified by OpenAI's o1 series. The ongoing progress in this field promises to reshape both our understanding of language and the application of AI in solving real-world problems.

\bibliographystyle{plain}

\begin{thebibliography}{100}

\bibitem{abdin2024phi}
Marah Abdin, Jyoti Aneja, Harkirat Behl, S{\'e}bastien Bubeck, Ronen Eldan, Suriya Gunasekar, Michael Harrison, Russell~J Hewett, Mojan Javaheripi, Piero Kauffmann, et~al.
\newblock Phi-4 technical report.
\newblock {\em arXiv preprint arXiv:2412.08905}, 2024.

\bibitem{achiam2023gpt}
Josh Achiam, Steven Adler, Sandhini Agarwal, Lama Ahmad, Ilge Akkaya, Florencia~Leoni Aleman, Diogo Almeida, Janko Altenschmidt, Sam Altman, Shyamal Anadkat, et~al.
\newblock Gpt-4 technical report.
\newblock {\em arXiv preprint arXiv:2303.08774}, 2023.

\bibitem{ahn2022can}
Michael Ahn, Anthony Brohan, Noah Brown, Yevgen Chebotar, Omar Cortes, Byron David, Chelsea Finn, Chuyuan Fu, Keerthana Gopalakrishnan, Karol Hausman, et~al.
\newblock Do as i can, not as i say: Grounding language in robotic affordances.
\newblock {\em arXiv preprint arXiv:2204.01691}, 2022.

\bibitem{amini2024direct}
Afra Amini, Tim Vieira, and Ryan Cotterell.
\newblock Direct preference optimization with an offset.
\newblock {\em arXiv preprint arXiv:2402.10571}, 2024.

\bibitem{amini2019mathqainterpretablemathword}
Aida Amini, Saadia Gabriel, Peter Lin, Rik Koncel-Kedziorski, Yejin Choi, and Hannaneh Hajishirzi.
\newblock Mathqa: Towards interpretable math word problem solving with operation-based formalisms, 2019.

\bibitem{austin2021program}
Jacob Austin, Augustus Odena, Maxwell Nye, Maarten Bosma, Henryk Michalewski, David Dohan, Ellen Jiang, Carrie Cai, Michael Terry, Quoc Le, et~al.
\newblock Program synthesis with large language models.
\newblock {\em arXiv preprint arXiv:2108.07732}, 2021.

\bibitem{openai2024o32}
AXIOS.
\newblock Openai's new o3 model freaks out computer science majors.
\newblock 2025.

\bibitem{bai2022training}
Yuntao Bai, Andy Jones, Kamal Ndousse, Amanda Askell, Anna Chen, Nova DasSarma, Dawn Drain, Stanislav Fort, Deep Ganguli, Tom Henighan, et~al.
\newblock Training a helpful and harmless assistant with reinforcement learning from human feedback.
\newblock {\em arXiv preprint arXiv:2204.05862}, 2022.

\bibitem{bai2022constitutional}
Yuntao Bai, Saurav Kadavath, Sandipan Kundu, Amanda Askell, Jackson Kernion, Andy Jones, Anna Chen, Anna Goldie, Azalia Mirhoseini, Cameron McKinnon, et~al.
\newblock Constitutional ai: Harmlessness from ai feedback.
\newblock {\em arXiv preprint arXiv:2212.08073}, 2022.

\bibitem{bakhtin2019phyre}
Anton Bakhtin, Laurens van~der Maaten, Justin Johnson, Laura Gustafson, and Ross Girshick.
\newblock Phyre: A new benchmark for physical reasoning.
\newblock {\em Advances in Neural Information Processing Systems}, 32, 2019.

\bibitem{besta2024graph}
Maciej Besta, Nils Blach, Ales Kubicek, Robert Gerstenberger, Michal Podstawski, Lukas Gianinazzi, Joanna Gajda, Tomasz Lehmann, Hubert Niewiadomski, Piotr Nyczyk, et~al.
\newblock Graph of thoughts: Solving elaborate problems with large language models.
\newblock In {\em Proceedings of the AAAI Conference on Artificial Intelligence}, volume~38, pages 17682--17690, 2024.

\bibitem{bisk2020piqa}
Yonatan Bisk, Rowan Zellers, Jianfeng Gao, Yejin Choi, et~al.
\newblock Piqa: Reasoning about physical commonsense in natural language.
\newblock In {\em Proceedings of the AAAI conference on artificial intelligence}, volume~34, pages 7432--7439, 2020.

\bibitem{boiko2023autonomous}
Daniil~A Boiko, Robert MacKnight, Ben Kline, and Gabe Gomes.
\newblock Autonomous chemical research with large language models.
\newblock {\em Nature}, 624(7992):570--578, 2023.

\bibitem{brown2020language}
Tom Brown, Benjamin Mann, Nick Ryder, Melanie Subbiah, Jared~D Kaplan, Prafulla Dhariwal, Arvind Neelakantan, Pranav Shyam, Girish Sastry, Amanda Askell, et~al.
\newblock Language models are few-shot learners.
\newblock {\em Advances in neural information processing systems}, 33:1877--1901, 2020.

\bibitem{browne2012survey}
Cameron~B Browne, Edward Powley, Daniel Whitehouse, Simon~M Lucas, Peter~I Cowling, Philipp Rohlfshagen, Stephen Tavener, Diego Perez, Spyridon Samothrakis, and Simon Colton.
\newblock A survey of monte carlo tree search methods.
\newblock {\em IEEE Transactions on Computational Intelligence and AI in games}, 4(1):1--43, 2012.

\bibitem{chen2024alphamath}
Guoxin Chen, Minpeng Liao, Chengxi Li, and Kai Fan.
\newblock Alphamath almost zero: process supervision without process.
\newblock {\em arXiv preprint arXiv:2405.03553}, 2024.

\bibitem{chen2024step}
Guoxin Chen, Minpeng Liao, Chengxi Li, and Kai Fan.
\newblock Step-level value preference optimization for mathematical reasoning.
\newblock {\em arXiv preprint arXiv:2406.10858}, 2024.

\bibitem{chen2022unigeounifyinggeometrylogical}
Jiaqi Chen, Tong Li, Jinghui Qin, Pan Lu, Liang Lin, Chongyu Chen, and Xiaodan Liang.
\newblock Unigeo: Unifying geometry logical reasoning via reformulating mathematical expression, 2022.

\bibitem{chen2022geoqageometricquestionanswering}
Jiaqi Chen, Jianheng Tang, Jinghui Qin, Xiaodan Liang, Lingbo Liu, Eric~P. Xing, and Liang Lin.
\newblock Geoqa: A geometric question answering benchmark towards multimodal numerical reasoning, 2022.

\bibitem{chen2024large}
Lin Chen, Fengli Xu, Nian Li, Zhenyu Han, Meng Wang, Yong Li, and Pan Hui.
\newblock Large language model-driven meta-structure discovery in heterogeneous information network.
\newblock In {\em Proceedings of the 30th ACM SIGKDD Conference on Knowledge Discovery and Data Mining}, pages 307--318, 2024.

\bibitem{chen2021evaluating}
Mark Chen, Jerry Tworek, Heewoo Jun, Qiming Yuan, Henrique Ponde De~Oliveira Pinto, Jared Kaplan, Harri Edwards, Yuri Burda, Nicholas Joseph, Greg Brockman, et~al.
\newblock Evaluating large language models trained on code.
\newblock {\em arXiv preprint arXiv:2107.03374}, 2021.

\bibitem{chen2023large}
Nuo Chen, Yan Wang, Haiyun Jiang, Deng Cai, Yuhan Li, Ziyang Chen, Longyue Wang, and Jia Li.
\newblock Large language models meet harry potter: A dataset for aligning dialogue agents with characters.
\newblock In {\em Findings of the Association for Computational Linguistics: EMNLP 2023}, pages 8506--8520, 2023.

\bibitem{chen2023theoremqatheoremdrivenquestionanswering}
Wenhu Chen, Ming Yin, Max Ku, Pan Lu, Yixin Wan, Xueguang Ma, Jianyu Xu, Xinyi Wang, and Tony Xia.
\newblock Theoremqa: A theorem-driven question answering dataset, 2023.

\bibitem{cobbe2021trainingverifierssolvemath}
Karl Cobbe, Vineet Kosaraju, Mohammad Bavarian, Mark Chen, Heewoo Jun, Lukasz Kaiser, Matthias Plappert, Jerry Tworek, Jacob Hilton, Reiichiro Nakano, Christopher Hesse, and John Schulman.
\newblock Training verifiers to solve math word problems, 2021.

\bibitem{coda2024meta}
Julian Coda-Forno, Marcel Binz, Zeynep Akata, Matt Botvinick, Jane Wang, and Eric Schulz.
\newblock Meta-in-context learning in large language models.
\newblock {\em Advances in Neural Information Processing Systems}, 36, 2024.

\bibitem{davis2024testing}
Ernest Davis.
\newblock Testing gpt-4-o1-preview on math and science problems: A follow-up study.
\newblock {\em arXiv preprint arXiv:2410.22340}, 2024.

\bibitem{de2024system}
Joost~CF de~Winter, Dimitra Dodou, and Yke~Bauke Eisma.
\newblock System 2 thinking in openai’s o1-preview model: Near-perfect performance on a mathematics exam.
\newblock {\em Computers}, 13(11):278, 2024.

\bibitem{deng2024mind2web}
Xiang Deng, Yu~Gu, Boyuan Zheng, Shijie Chen, Sam Stevens, Boshi Wang, Huan Sun, and Yu~Su.
\newblock Mind2web: Towards a generalist agent for the web.
\newblock {\em Advances in Neural Information Processing Systems}, 36, 2024.

\bibitem{deng2021reasonbert}
Xiang Deng, Yu~Su, Alyssa Lees, You Wu, Cong Yu, and Huan Sun.
\newblock Reasonbert: Pre-trained to reason with distant supervision.
\newblock {\em arXiv preprint arXiv:2109.04912}, 2021.

\bibitem{devlin2018bert}
Jacob Devlin.
\newblock Bert: Pre-training of deep bidirectional transformers for language understanding.
\newblock {\em arXiv preprint arXiv:1810.04805}, 2018.

\bibitem{ding2024data}
Bosheng Ding, Chengwei Qin, Ruochen Zhao, Tianze Luo, Xinze Li, Guizhen Chen, Wenhan Xia, Junjie Hu, Anh~Tuan Luu, and Shafiq Joty.
\newblock Data augmentation using llms: Data perspectives, learning paradigms and challenges.
\newblock {\em arXiv preprint arXiv:2403.02990}, 2024.

\bibitem{ding2023enhancing}
Ning Ding, Yulin Chen, Bokai Xu, Yujia Qin, Zhi Zheng, Shengding Hu, Zhiyuan Liu, Maosong Sun, and Bowen Zhou.
\newblock Enhancing chat language models by scaling high-quality instructional conversations.
\newblock {\em arXiv preprint arXiv:2305.14233}, 2023.

\bibitem{dong2022survey}
Qingxiu Dong, Lei Li, Damai Dai, Ce~Zheng, Zhiyong Wu, Baobao Chang, Xu~Sun, Jingjing Xu, and Zhifang Sui.
\newblock A survey on in-context learning.
\newblock {\em arXiv preprint arXiv:2301.00234}, 2022.

\bibitem{sep-abduction}
Igor Douven.
\newblock {Abduction}.
\newblock In Edward~N. Zalta, editor, {\em The {Stanford} Encyclopedia of Philosophy}. Metaphysics Research Lab, Stanford University, {S}ummer 2021 edition, 2021.

\bibitem{dubey2024llama}
Abhimanyu Dubey, Abhinav Jauhri, Abhinav Pandey, Abhishek Kadian, Ahmad Al-Dahle, Aiesha Letman, Akhil Mathur, Alan Schelten, Amy Yang, Angela Fan, et~al.
\newblock The llama 3 herd of models.
\newblock {\em arXiv preprint arXiv:2407.21783}, 2024.

\bibitem{duenas2024path}
Tom Duenas and Diana Ruiz.
\newblock The path to superintelligence: A critical analysis of openai’s five levels of ai progression.
\newblock {\em ResearchGate, 2024b. doi}, 10, 2024.

\bibitem{farquhar2024detecting}
Sebastian Farquhar, Jannik Kossen, Lorenz Kuhn, and Yarin Gal.
\newblock Detecting hallucinations in large language models using semantic entropy.
\newblock {\em Nature}, 630(8017):625--630, 2024.

\bibitem{fernando2023promptbreeder}
Chrisantha Fernando, Dylan Banarse, Henryk Michalewski, Simon Osindero, and Tim Rockt{\"a}schel.
\newblock Promptbreeder: Self-referential self-improvement via prompt evolution.
\newblock {\em arXiv preprint arXiv:2309.16797}, 2023.

\bibitem{gao2024memory}
Hang Gao and Yongfeng Zhang.
\newblock Memory sharing for large language model based agents.
\newblock {\em arXiv preprint arXiv:2404.09982}, 2024.

\bibitem{gao2024self}
Jinglong Gao, Xiao Ding, Yiming Cui, Jianbai Zhao, Hepeng Wang, Ting Liu, and Bing Qin.
\newblock Self-evolving gpt: A lifelong autonomous experiential learner.
\newblock {\em arXiv preprint arXiv:2407.08937}, 2024.

\bibitem{gehring2024rlef}
Jonas Gehring, Kunhao Zheng, Jade Copet, Vegard Mella, Taco Cohen, and Gabriel Synnaeve.
\newblock Rlef: Grounding code llms in execution feedback with reinforcement learning.
\newblock {\em arXiv preprint arXiv:2410.02089}, 2024.

\bibitem{goel2023llms}
Akshay Goel, Almog Gueta, Omry Gilon, Chang Liu, Sofia Erell, Lan~Huong Nguyen, Xiaohong Hao, Bolous Jaber, Shashir Reddy, Rupesh Kartha, et~al.
\newblock Llms accelerate annotation for medical information extraction.
\newblock In {\em Machine Learning for Health (ML4H)}, pages 82--100. PMLR, 2023.

\bibitem{gou2023critic}
Zhibin Gou, Zhihong Shao, Yeyun Gong, Yelong Shen, Yujiu Yang, Nan Duan, and Weizhu Chen.
\newblock Critic: Large language models can self-correct with tool-interactive critiquing.
\newblock {\em arXiv preprint arXiv:2305.11738}, 2023.

\bibitem{guan2024richelieu}
Zhenyu Guan, Xiangyu Kong, Fangwei Zhong, and Yizhou Wang.
\newblock Richelieu: Self-evolving llm-based agents for ai diplomacy.
\newblock {\em arXiv preprint arXiv:2407.06813}, 2024.

\bibitem{gulcehre2023reinforced}
Caglar Gulcehre, Tom~Le Paine, Srivatsan Srinivasan, Ksenia Konyushkova, Lotte Weerts, Abhishek Sharma, Aditya Siddhant, Alex Ahern, Miaosen Wang, Chenjie Gu, et~al.
\newblock Reinforced self-training (rest) for language modeling.
\newblock {\em arXiv preprint arXiv:2308.08998}, 2023.

\bibitem{han2024folionaturallanguagereasoning}
Simeng Han, Hailey Schoelkopf, Yilun Zhao, Zhenting Qi, Martin Riddell, Wenfei Zhou, James Coady, David Peng, Yujie Qiao, Luke Benson, Lucy Sun, Alex Wardle-Solano, Hannah Szabo, Ekaterina Zubova, Matthew Burtell, Jonathan Fan, Yixin Liu, Brian Wong, Malcolm Sailor, Ansong Ni, Linyong Nan, Jungo Kasai, Tao Yu, Rui Zhang, Alexander~R. Fabbri, Wojciech Kryscinski, Semih Yavuz, Ye~Liu, Xi~Victoria Lin, Shafiq Joty, Yingbo Zhou, Caiming Xiong, Rex Ying, Arman Cohan, and Dragomir Radev.
\newblock Folio: Natural language reasoning with first-order logic, 2024.

\bibitem{sep-logic-inductive}
James Hawthorne.
\newblock {Inductive Logic}.
\newblock In Edward~N. Zalta and Uri Nodelman, editors, {\em The {Stanford} Encyclopedia of Philosophy}. Metaphysics Research Lab, Stanford University, {S}ummer 2024 edition, 2024.

\bibitem{hayawi2024cross}
Kadhim Hayawi and Sakib Shahriar.
\newblock A cross-domain performance report of open ai chatgpt o1 model.
\newblock 2024.

\bibitem{hendrycks2021measuring}
Dan Hendrycks, Steven Basart, Saurav Kadavath, Mantas Mazeika, Akul Arora, Ethan Guo, Collin Burns, Samir Puranik, Horace He, Dawn Song, et~al.
\newblock Measuring coding challenge competence with apps.
\newblock {\em arXiv preprint arXiv:2105.09938}, 2021.

\bibitem{hendrycks2021measuringmathematicalproblemsolving}
Dan Hendrycks, Collin Burns, Saurav Kadavath, Akul Arora, Steven Basart, Eric Tang, Dawn Song, and Jacob Steinhardt.
\newblock Measuring mathematical problem solving with the math dataset, 2021.

\bibitem{hosseini-etal-2014-learning}
Mohammad~Javad Hosseini, Hannaneh Hajishirzi, Oren Etzioni, and Nate Kushman.
\newblock Learning to solve arithmetic word problems with verb categorization.
\newblock In Alessandro Moschitti, Bo~Pang, and Walter Daelemans, editors, {\em Proceedings of the 2014 Conference on Empirical Methods in Natural Language Processing ({EMNLP})}, pages 523--533, Doha, Qatar, October 2014. Association for Computational Linguistics.

\bibitem{hu2024can}
Haichuan Hu, Ye~Shang, Guolin Xu, Congqing He, and Quanjun Zhang.
\newblock Can gpt-o1 kill all bugs? an evaluation of gpt-family llms on quixbugs.
\newblock {\em arXiv e-prints}, pages arXiv--2409, 2024.

\bibitem{hu2024automated}
Shengran Hu, Cong Lu, and Jeff Clune.
\newblock Automated design of agentic systems.
\newblock {\em arXiv preprint arXiv:2408.08435}, 2024.

\bibitem{huot2024agents}
Fantine Huot, Reinald~Kim Amplayo, Jennimaria Palomaki, Alice~Shoshana Jakobovits, Elizabeth Clark, and Mirella Lapata.
\newblock Agents' room: Narrative generation through multi-step collaboration.
\newblock {\em arXiv preprint arXiv:2410.02603}, 2024.

\bibitem{hwang2024self}
Hyeonbin Hwang, Doyoung Kim, Seungone Kim, Seonghyeon Ye, and Minjoon Seo.
\newblock Self-explore to avoid the pit: Improving the reasoning capabilities of language models with fine-grained rewards.
\newblock {\em arXiv preprint arXiv:2404.10346}, 2024.

\bibitem{blob-RFT}
interconnects.ai.
\newblock blob reinforcement fine-tuning.
\newblock (Accessed: 2025-12-6).

\bibitem{jiang2021lisa}
Albert~Q. Jiang, Wenda Li, Jesse~Michael Han, and Yuhuai Wu.
\newblock Lisa: Language models of isabelle proofs.
\newblock {\em 6th Conference on Artificial Intelligence and Theorem Proving}, 2021.

\bibitem{jimenez2023swe}
Carlos~E Jimenez, John Yang, Alexander Wettig, Shunyu Yao, Kexin Pei, Ofir Press, and Karthik Narasimhan.
\newblock Swe-bench: Can language models resolve real-world github issues?
\newblock {\em arXiv preprint arXiv:2310.06770}, 2023.

\bibitem{kaplan2020scaling}
Jared Kaplan, Sam McCandlish, Tom Henighan, Tom~B Brown, Benjamin Chess, Rewon Child, Scott Gray, Alec Radford, Jeffrey Wu, and Dario Amodei.
\newblock Scaling laws for neural language models.
\newblock {\em arXiv preprint arXiv:2001.08361}, 2020.

\bibitem{kazemnejad2024vineppo}
Amirhossein Kazemnejad, Milad Aghajohari, Eva Portelance, Alessandro Sordoni, Siva Reddy, Aaron Courville, and Nicolas~Le Roux.
\newblock Vineppo: Unlocking rl potential for llm reasoning through refined credit assignment.
\newblock {\em arXiv preprint arXiv:2410.01679}, 2024.

\bibitem{kim2024meganno+}
Hannah Kim, Kushan Mitra, Rafael~Li Chen, Sajjadur Rahman, and Dan Zhang.
\newblock Meganno+: A human-llm collaborative annotation system.
\newblock {\em arXiv preprint arXiv:2402.18050}, 2024.

\bibitem{kojima2022large}
Takeshi Kojima, Shixiang~Shane Gu, Machel Reid, Yutaka Matsuo, and Yusuke Iwasawa.
\newblock Large language models are zero-shot reasoners.
\newblock {\em Advances in neural information processing systems}, 35:22199--22213, 2022.

\bibitem{kondo2023probing}
Kazushi Kondo, Saku Sugawara, and Akiko Aizawa.
\newblock Probing physical reasoning with counter-commonsense context.
\newblock In {\em Proceedings of the 61st Annual Meeting of the Association for Computational Linguistics (Volume 2: Short Papers)}, pages 603--612, 2023.

\bibitem{kumar2024training}
Aviral Kumar, Vincent Zhuang, Rishabh Agarwal, Yi~Su, John~D Co-Reyes, Avi Singh, Kate Baumli, Shariq Iqbal, Colton Bishop, Rebecca Roelofs, et~al.
\newblock Training language models to self-correct via reinforcement learning.
\newblock {\em arXiv preprint arXiv:2409.12917}, 2024.

\bibitem{kwon2024language}
Teyun Kwon, Norman Di~Palo, and Edward Johns.
\newblock Language models as zero-shot trajectory generators.
\newblock {\em IEEE Robotics and Automation Letters}, 2024.

\bibitem{laban2023llms}
Philippe Laban, Wojciech Kry{\'s}ci{\'n}ski, Divyansh Agarwal, Alexander~R Fabbri, Caiming Xiong, Shafiq Joty, and Chien-Sheng Wu.
\newblock Llms as factual reasoners: Insights from existing benchmarks and beyond.
\newblock {\em arXiv preprint arXiv:2305.14540}, 2023.

\bibitem{lai2023ds}
Yuhang Lai, Chengxi Li, Yiming Wang, Tianyi Zhang, Ruiqi Zhong, Luke Zettlemoyer, Wen-tau Yih, Daniel Fried, Sida Wang, and Tao Yu.
\newblock Ds-1000: A natural and reliable benchmark for data science code generation.
\newblock In {\em International Conference on Machine Learning}, pages 18319--18345. PMLR, 2023.

\bibitem{latif2024systematic}
Ehsan Latif, Yifan Zhou, Shuchen Guo, Yizhu Gao, Lehong Shi, Matthew Nayaaba, Gyeonggeon Lee, Liang Zhang, Arne Bewersdorff, Luyang Fang, et~al.
\newblock A systematic assessment of openai o1-preview for higher order thinking in education.
\newblock {\em arXiv preprint arXiv:2410.21287}, 2024.

\bibitem{le2024multi}
Hao~Duong Le, Xin Xia, and Zhang Chen.
\newblock Multi-agent causal discovery using large language models.
\newblock {\em arXiv preprint arXiv:2407.15073}, 2024.

\bibitem{lee2024llm2llm}
Nicholas Lee, Thanakul Wattanawong, Sehoon Kim, Karttikeya Mangalam, Sheng Shen, Gopala Anumanchipalli, Michael~W Mahoney, Kurt Keutzer, and Amir Gholami.
\newblock Llm2llm: Boosting llms with novel iterative data enhancement.
\newblock {\em arXiv preprint arXiv:2403.15042}, 2024.

\bibitem{lewis2019bart}
M~Lewis.
\newblock Bart: Denoising sequence-to-sequence pre-training for natural language generation, translation, and comprehension.
\newblock {\em arXiv preprint arXiv:1910.13461}, 2019.

\bibitem{li2024dotamath}
Chengpeng Li, Guanting Dong, Mingfeng Xue, Ru~Peng, Xiang Wang, and Dayiheng Liu.
\newblock Dotamath: Decomposition of thought with code assistance and self-correction for mathematical reasoning.
\newblock {\em arXiv preprint arXiv:2407.04078}, 2024.

\bibitem{li2024openai}
Leo Li, Ye~Luo, and Tingyou Pan.
\newblock Openai-o1 ab testing: Does the o1 model really do good reasoning in math problem solving?
\newblock {\em arXiv preprint arXiv:2411.06198}, 2024.

\bibitem{li2023coannotating}
Minzhi Li, Taiwei Shi, Caleb Ziems, Min-Yen Kan, Nancy~F Chen, Zhengyuan Liu, and Diyi Yang.
\newblock Coannotating: Uncertainty-guided work allocation between human and large language models for data annotation.
\newblock {\em arXiv preprint arXiv:2310.15638}, 2023.

\bibitem{lightman2023let}
Hunter Lightman, Vineet Kosaraju, Yura Burda, Harri Edwards, Bowen Baker, Teddy Lee, Jan Leike, John Schulman, Ilya Sutskever, and Karl Cobbe.
\newblock Let's verify step by step.
\newblock {\em arXiv preprint arXiv:2305.20050}, 2023.

\bibitem{lightman2023letsverifystepstep}
Hunter Lightman, Vineet Kosaraju, Yura Burda, Harri Edwards, Bowen Baker, Teddy Lee, Jan Leike, John Schulman, Ilya Sutskever, and Karl Cobbe.
\newblock Let's verify step by step, 2023.

\bibitem{liu2022wanliworkeraicollaboration}
Alisa Liu, Swabha Swayamdipta, Noah~A. Smith, and Yejin Choi.
\newblock Wanli: Worker and ai collaboration for natural language inference dataset creation, 2022.

\bibitem{liu2023fimochallengeformaldataset}
Chengwu Liu, Jianhao Shen, Huajian Xin, Zhengying Liu, Ye~Yuan, Haiming Wang, Wei Ju, Chuanyang Zheng, Yichun Yin, Lin Li, Ming Zhang, and Qun Liu.
\newblock Fimo: A challenge formal dataset for automated theorem proving, 2023.

\bibitem{liu2023agentbench}
Xiao Liu, Hao Yu, Hanchen Zhang, Yifan Xu, Xuanyu Lei, Hanyu Lai, Yu~Gu, Hangliang Ding, Kaiwen Men, Kejuan Yang, et~al.
\newblock Agentbench: Evaluating llms as agents.
\newblock {\em arXiv preprint arXiv:2308.03688}, 2023.

\bibitem{lobo2024impact}
Elita Lobo, Chirag Agarwal, and Himabindu Lakkaraju.
\newblock On the impact of fine-tuning on chain-of-thought reasoning.
\newblock {\em arXiv preprint arXiv:2411.15382}, 2024.

\bibitem{lu2024mathvistaevaluatingmathematicalreasoning}
Pan Lu, Hritik Bansal, Tony Xia, Jiacheng Liu, Chunyuan Li, Hannaneh Hajishirzi, Hao Cheng, Kai-Wei Chang, Michel Galley, and Jianfeng Gao.
\newblock Mathvista: Evaluating mathematical reasoning of foundation models in visual contexts, 2024.

\bibitem{lu2021intergpsinterpretablegeometryproblem}
Pan Lu, Ran Gong, Shibiao Jiang, Liang Qiu, Siyuan Huang, Xiaodan Liang, and Song-Chun Zhu.
\newblock Inter-gps: Interpretable geometry problem solving with formal language and symbolic reasoning, 2021.

\bibitem{lu2023dynamicpromptlearningpolicy}
Pan Lu, Liang Qiu, Kai-Wei Chang, Ying~Nian Wu, Song-Chun Zhu, Tanmay Rajpurohit, Peter Clark, and Ashwin Kalyan.
\newblock Dynamic prompt learning via policy gradient for semi-structured mathematical reasoning, 2023.

\bibitem{luo2023wizardmath}
Haipeng Luo, Qingfeng Sun, Can Xu, Pu~Zhao, Jianguang Lou, Chongyang Tao, Xiubo Geng, Qingwei Lin, Shifeng Chen, and Dongmei Zhang.
\newblock Wizardmath: Empowering mathematical reasoning for large language models via reinforced evol-instruct.
\newblock {\em arXiv preprint arXiv:2308.09583}, 2023.

\bibitem{luo2024improve}
Liangchen Luo, Yinxiao Liu, Rosanne Liu, Samrat Phatale, Harsh Lara, Yunxuan Li, Lei Shu, Yun Zhu, Lei Meng, Jiao Sun, et~al.
\newblock Improve mathematical reasoning in language models by automated process supervision.
\newblock {\em arXiv preprint arXiv:2406.06592}, 2024.

\bibitem{luo2024logigluebriefsurveybenchmark}
Man Luo, Shrinidhi Kumbhar, Ming shen, Mihir Parmar, Neeraj Varshney, Pratyay Banerjee, Somak Aditya, and Chitta Baral.
\newblock Towards logiglue: A brief survey and a benchmark for analyzing logical reasoning capabilities of language models, 2024.

\bibitem{ma2024agentboard}
Chang Ma, Junlei Zhang, Zhihao Zhu, Cheng Yang, Yujiu Yang, Yaohui Jin, Zhenzhong Lan, Lingpeng Kong, and Junxian He.
\newblock Agentboard: An analytical evaluation board of multi-turn llm agents.
\newblock {\em arXiv preprint arXiv:2401.13178}, 2024.

\bibitem{ma2024llm}
Pingchuan Ma, Tsun-Hsuan Wang, Minghao Guo, Zhiqing Sun, Joshua~B Tenenbaum, Daniela Rus, Chuang Gan, and Wojciech Matusik.
\newblock Llm and simulation as bilevel optimizers: A new paradigm to advance physical scientific discovery.
\newblock {\em arXiv preprint arXiv:2405.09783}, 2024.

\bibitem{maas2023infinity}
Carey Maas, Saatchi Wheeler, Shamash Billington, et~al.
\newblock To infinity and beyond: Show-1 and showrunner agents in multi-agent simulations.
\newblock {\em To infinity and beyond: Show-1 and showrunner agents in multi-agent simulations}, 2023.

\bibitem{madaan2024self}
Aman Madaan, Niket Tandon, Prakhar Gupta, Skyler Hallinan, Luyu Gao, Sarah Wiegreffe, Uri Alon, Nouha Dziri, Shrimai Prabhumoye, Yiming Yang, et~al.
\newblock Self-refine: Iterative refinement with self-feedback.
\newblock {\em Advances in Neural Information Processing Systems}, 36, 2024.

\bibitem{mall2023remote}
Utkarsh Mall, Cheng~Perng Phoo, Meilin~Kelsey Liu, Carl Vondrick, Bharath Hariharan, and Kavita Bala.
\newblock Remote sensing vision-language foundation models without annotations via ground remote alignment.
\newblock {\em arXiv preprint arXiv:2312.06960}, 2023.

\bibitem{mao2024champcompetitionleveldatasetfinegrained}
Yujun Mao, Yoon Kim, and Yilun Zhou.
\newblock Champ: A competition-level dataset for fine-grained analyses of llms' mathematical reasoning capabilities, 2024.

\bibitem{masry2022chartqabenchmarkquestionanswering}
Ahmed Masry, Do~Xuan Long, Jia~Qing Tan, Shafiq Joty, and Enamul Hoque.
\newblock Chartqa: A benchmark for question answering about charts with visual and logical reasoning, 2022.

\bibitem{mccoy2024language}
R~Thomas McCoy, Shunyu Yao, Dan Friedman, Mathew~D Hardy, and Thomas~L Griffiths.
\newblock When a language model is optimized for reasoning, does it still show embers of autoregression? an analysis of openai o1.
\newblock {\em arXiv preprint arXiv:2410.01792}, 2024.

\bibitem{mihaylov2018can}
Todor Mihaylov, Peter Clark, Tushar Khot, and Ashish Sabharwal.
\newblock Can a suit of armor conduct electricity? a new dataset for open book question answering.
\newblock In {\em Proceedings of the 2018 Conference on Empirical Methods in Natural Language Processing}, pages 2381--2391, 2018.

\bibitem{mikulova2023quality}
Marie Mikulov{\'a}, Milan Straka, Jan {\v{S}}t{\v{e}}p{\'a}nek, Barbora {\v{S}}t{\v{e}}p{\'a}nkov{\'a}, and Jan Haji{\v{c}}.
\newblock Quality and efficiency of manual annotation: Pre-annotation bias.
\newblock {\em arXiv preprint arXiv:2306.09307}, 2023.

\bibitem{min2022rethinking}
Sewon Min, Xinxi Lyu, Ari Holtzman, Mikel Artetxe, Mike Lewis, Hannaneh Hajishirzi, and Luke Zettlemoyer.
\newblock Rethinking the role of demonstrations: What makes in-context learning work?
\newblock {\em arXiv preprint arXiv:2202.12837}, 2022.

\bibitem{nasution2024chatgpt}
Arbi~Haza Nasution and Aytug Onan.
\newblock Chatgpt label: Comparing the quality of human-generated and llm-generated annotations in low-resource language nlp tasks.
\newblock {\em IEEE Access}, 2024.

\bibitem{nie2020adversarialnlinewbenchmark}
Yixin Nie, Adina Williams, Emily Dinan, Mohit Bansal, Jason Weston, and Douwe Kiela.
\newblock Adversarial nli: A new benchmark for natural language understanding, 2020.

\bibitem{niven2019probingneuralnetworkcomprehension}
Timothy Niven and Hung-Yu Kao.
\newblock Probing neural network comprehension of natural language arguments, 2019.

\bibitem{openai-RFT}
OpenAI.
\newblock Reinforcement fine-tuning.
\newblock (Accessed: 2025-12-6).

\bibitem{openai2024o3}
OpenAI.
\newblock Early access for safety testing.
\newblock 2024.

\bibitem{openai2024learning}
OpenAI.
\newblock Learning to reason with llms.
\newblock 2024.

\bibitem{ouyang2022training}
Long Ouyang, Jeffrey Wu, Xu~Jiang, Diogo Almeida, Carroll Wainwright, Pamela Mishkin, Chong Zhang, Sandhini Agarwal, Katarina Slama, Alex Ray, et~al.
\newblock Training language models to follow instructions with human feedback.
\newblock {\em Advances in neural information processing systems}, 35:27730--27744, 2022.

\bibitem{park2023generative}
Joon~Sung Park, Joseph O'Brien, Carrie~Jun Cai, Meredith~Ringel Morris, Percy Liang, and Michael~S Bernstein.
\newblock Generative agents: Interactive simulacra of human behavior.
\newblock In {\em Proceedings of the 36th Annual ACM Symposium on User Interface Software and Technology}, pages 1--22, 2023.

\bibitem{puri2020training}
Raul Puri, Ryan Spring, Mostofa Patwary, Mohammad Shoeybi, and Bryan Catanzaro.
\newblock Training question answering models from synthetic data.
\newblock {\em arXiv preprint arXiv:2002.09599}, 2020.

\bibitem{qian2023communicative}
Chen Qian, Xin Cong, Cheng Yang, Weize Chen, Yusheng Su, Juyuan Xu, Zhiyuan Liu, and Maosong Sun.
\newblock Communicative agents for software development.
\newblock {\em arXiv preprint arXiv:2307.07924}, 2023.

\bibitem{qian2023experiential}
Chen Qian, Yufan Dang, Jiahao Li, Wei Liu, Zihao Xie, Yifei Wang, Weize Chen, Cheng Yang, Xin Cong, Xiaoyin Che, et~al.
\newblock Experiential co-learning of software-developing agents.
\newblock {\em arXiv preprint arXiv:2312.17025}, 2023.

\bibitem{qiao2024autoact}
Shuofei Qiao, Ningyu Zhang, Runnan Fang, Yujie Luo, Wangchunshu Zhou, Yuchen~Eleanor Jiang, Chengfei Lv, and Huajun Chen.
\newblock Autoact: Automatic agent learning from scratch via self-planning.
\newblock {\em arXiv preprint arXiv:2401.05268}, 2024.

\bibitem{qin2024o1}
Yiwei Qin, Xuefeng Li, Haoyang Zou, Yixiu Liu, Shijie Xia, Zhen Huang, Yixin Ye, Weizhe Yuan, Hector Liu, Yuanzhi Li, et~al.
\newblock O1 replication journey: A strategic progress report--part 1.
\newblock {\em arXiv preprint arXiv:2410.18982}, 2024.

\bibitem{radford2018improving}
Alec Radford.
\newblock Improving language understanding by generative pre-training.
\newblock 2018.

\bibitem{rafailov2024direct}
Rafael Rafailov, Archit Sharma, Eric Mitchell, Christopher~D Manning, Stefano Ermon, and Chelsea Finn.
\newblock Direct preference optimization: Your language model is secretly a reward model.
\newblock {\em Advances in Neural Information Processing Systems}, 36, 2024.

\bibitem{raffel2020exploring}
Colin Raffel, Noam Shazeer, Adam Roberts, Katherine Lee, Sharan Narang, Michael Matena, Yanqi Zhou, Wei Li, and Peter~J Liu.
\newblock Exploring the limits of transfer learning with a unified text-to-text transformer.
\newblock {\em Journal of machine learning research}, 21(140):1--67, 2020.

\bibitem{ramamurthy2022reinforcement}
Rajkumar Ramamurthy, Prithviraj Ammanabrolu, Kiant{\'e} Brantley, Jack Hessel, Rafet Sifa, Christian Bauckhage, Hannaneh Hajishirzi, and Yejin Choi.
\newblock Is reinforcement learning (not) for natural language processing: Benchmarks, baselines, and building blocks for natural language policy optimization.
\newblock {\em arXiv preprint arXiv:2210.01241}, 2022.

\bibitem{romera2024mathematical}
Bernardino Romera-Paredes, Mohammadamin Barekatain, Alexander Novikov, Matej Balog, M~Pawan Kumar, Emilien Dupont, Francisco~JR Ruiz, Jordan~S Ellenberg, Pengming Wang, Omar Fawzi, et~al.
\newblock Mathematical discoveries from program search with large language models.
\newblock {\em Nature}, 625(7995):468--475, 2024.

\bibitem{roy2016solvinggeneralarithmeticword}
Subhro Roy and Dan Roth.
\newblock Solving general arithmetic word problems, 2016.

\bibitem{sakaguchi2020winogrande}
Keisuke Sakaguchi, Ronan Le~Bras, Chandra Bhagavatula, and Yejin Choi.
\newblock Winogrande: An adversarial winograd schema challenge at scale.
\newblock In {\em Proceedings of the AAAI Conference on Artificial Intelligence}, volume~34, pages 8732--8740, 2020.

\bibitem{sap2019social}
Maarten Sap, Hannah Rashkin, Derek Chen, Ronan Le~Bras, and Yejin Choi.
\newblock Social iqa: Commonsense reasoning about social interactions.
\newblock In {\em Proceedings of the 2019 Conference on Empirical Methods in Natural Language Processing and the 9th International Joint Conference on Natural Language Processing (EMNLP-IJCNLP)}, pages 4463--4473, 2019.

\bibitem{saparov2023languagemodelsgreedyreasoners}
Abulhair Saparov and He~He.
\newblock Language models are greedy reasoners: A systematic formal analysis of chain-of-thought, 2023.

\bibitem{schick2024toolformer}
Timo Schick, Jane Dwivedi-Yu, Roberto Dess{\`\i}, Roberta Raileanu, Maria Lomeli, Eric Hambro, Luke Zettlemoyer, Nicola Cancedda, and Thomas Scialom.
\newblock Toolformer: Language models can teach themselves to use tools.
\newblock {\em Advances in Neural Information Processing Systems}, 36, 2024.

\bibitem{schulman2017proximal}
John Schulman, Filip Wolski, Prafulla Dhariwal, Alec Radford, and Oleg Klimov.
\newblock Proximal policy optimization algorithms.
\newblock {\em arXiv preprint arXiv:1707.06347}, 2017.

\bibitem{sel2023algorithm}
Bilgehan Sel, Ahmad Al-Tawaha, Vanshaj Khattar, Ruoxi Jia, and Ming Jin.
\newblock Algorithm of thoughts: Enhancing exploration of ideas in large language models.
\newblock {\em arXiv preprint arXiv:2308.10379}, 2023.

\bibitem{setlur2024rewarding}
Amrith Setlur, Chirag Nagpal, Adam Fisch, Xinyang Geng, Jacob Eisenstein, Rishabh Agarwal, Alekh Agarwal, Jonathan Berant, and Aviral Kumar.
\newblock Rewarding progress: Scaling automated process verifiers for llm reasoning.
\newblock {\em arXiv preprint arXiv:2410.08146}, 2024.

\bibitem{shanahan2023role}
Murray Shanahan, Kyle McDonell, and Laria Reynolds.
\newblock Role play with large language models.
\newblock {\em Nature}, 623(7987):493--498, 2023.

\bibitem{shang2024agentsquare}
Yu~Shang, Yu~Li, Keyu Zhao, Likai Ma, Jiahe Liu, Fengli Xu, and Yong Li.
\newblock Agentsquare: Automatic llm agent search in modular design space.
\newblock {\em arXiv preprint arXiv:2410.06153}, 2024.

\bibitem{shani2024multi}
Lior Shani, Aviv Rosenberg, Asaf Cassel, Oran Lang, Daniele Calandriello, Avital Zipori, Hila Noga, Orgad Keller, Bilal Piot, Idan Szpektor, et~al.
\newblock Multi-turn reinforcement learning from preference human feedback.
\newblock {\em arXiv preprint arXiv:2405.14655}, 2024.

\bibitem{shao2024beyond}
Chenyang Shao, Fengli Xu, Bingbing Fan, Jingtao Ding, Yuan Yuan, Meng Wang, and Yong Li.
\newblock Beyond imitation: Generating human mobility from context-aware reasoning with large language models.
\newblock {\em arXiv preprint arXiv:2402.09836}, 2024.

\bibitem{shao2024deepseekmath}
Zhihong Shao, Peiyi Wang, Qihao Zhu, Runxin Xu, Junxiao Song, Xiao Bi, Haowei Zhang, Mingchuan Zhang, YK~Li, Y~Wu, et~al.
\newblock Deepseekmath: Pushing the limits of mathematical reasoning in open language models.
\newblock {\em arXiv preprint arXiv:2402.03300}, 2024.

\bibitem{shinn2024reflexion}
Noah Shinn, Federico Cassano, Ashwin Gopinath, Karthik Narasimhan, and Shunyu Yao.
\newblock Reflexion: Language agents with verbal reinforcement learning.
\newblock {\em Advances in Neural Information Processing Systems}, 36, 2024.

\bibitem{shojaee2024llm}
Parshin Shojaee, Kazem Meidani, Shashank Gupta, Amir~Barati Farimani, and Chandan~K Reddy.
\newblock Llm-sr: Scientific equation discovery via programming with large language models.
\newblock {\em arXiv preprint arXiv:2404.18400}, 2024.

\bibitem{shridhar2020alfworld}
Mohit Shridhar, Xingdi Yuan, Marc-Alexandre C{\^o}t{\'e}, Yonatan Bisk, Adam Trischler, and Matthew Hausknecht.
\newblock Alfworld: Aligning text and embodied environments for interactive learning.
\newblock {\em arXiv preprint arXiv:2010.03768}, 2020.

\bibitem{sinha2019clutrrdiagnosticbenchmarkinductive}
Koustuv Sinha, Shagun Sodhani, Jin Dong, Joelle Pineau, and William~L. Hamilton.
\newblock Clutrr: A diagnostic benchmark for inductive reasoning from text, 2019.

\bibitem{smoke1961program}
W~Smoke and E~Dubinsky.
\newblock A program for the machine translation of natural languages.
\newblock {\em Mech. Transl. Comput. Linguistics}, 6:2--10, 1961.

\bibitem{snell2024scaling}
Charlie Snell, Jaehoon Lee, Kelvin Xu, and Aviral Kumar.
\newblock Scaling llm test-time compute optimally can be more effective than scaling model parameters.
\newblock {\em arXiv preprint arXiv:2408.03314}, 2024.

\bibitem{song2024trial}
Yifan Song, Da~Yin, Xiang Yue, Jie Huang, Sujian Li, and Bill~Yuchen Lin.
\newblock Trial and error: Exploration-based trajectory optimization for llm agents.
\newblock {\em arXiv preprint arXiv:2403.02502}, 2024.

\bibitem{srivastava2022beyond}
Aarohi Srivastava, Abhinav Rastogi, Abhishek Rao, Abu Awal~Md Shoeb, Abubakar Abid, Adam Fisch, Adam~R Brown, Adam Santoro, Aditya Gupta, Adri{\`a} Garriga-Alonso, et~al.
\newblock Beyond the imitation game: Quantifying and extrapolating the capabilities of language models.
\newblock {\em arXiv preprint arXiv:2206.04615}, 2022.

\bibitem{sumers2023cognitive}
Theodore~R Sumers, Shunyu Yao, Karthik Narasimhan, and Thomas~L Griffiths.
\newblock Cognitive architectures for language agents.
\newblock {\em arXiv preprint arXiv:2309.02427}, 2023.

\bibitem{sun2024retrievalaugmentedhierarchicalincontextreinforcement}
Chuanneng Sun, Songjun Huang, and Dario Pompili.
\newblock Retrieval-augmented hierarchical in-context reinforcement learning and hindsight modular reflections for task planning with llms, 2024.

\bibitem{sutton2019bitter}
Richard Sutton.
\newblock The bitter lesson.
\newblock {\em Incomplete Ideas (blog)}, 13(1):38, 2019.

\bibitem{tafjord2021proofwritergeneratingimplicationsproofs}
Oyvind Tafjord, Bhavana~Dalvi Mishra, and Peter Clark.
\newblock Proofwriter: Generating implications, proofs, and abductive statements over natural language, 2021.

\bibitem{talmor2019commonsenseqa}
Alon Talmor, Jonathan Herzig, Nicholas Lourie, and Jonathan Berant.
\newblock Commonsenseqa: A question answering challenge targeting commonsense knowledge.
\newblock In {\em Proceedings of the 2019 Conference of the North American Chapter of the Association for Computational Linguistics: Human Language Technologies, Volume 1 (Long and Short Papers)}, pages 4149--4158, 2019.

\bibitem{tan2024large}
Zhen Tan, Dawei Li, Song Wang, Alimohammad Beigi, Bohan Jiang, Amrita Bhattacharjee, Mansooreh Karami, Jundong Li, Lu~Cheng, and Huan Liu.
\newblock Large language models for data annotation: A survey.
\newblock {\em arXiv preprint arXiv:2402.13446}, 2024.

\bibitem{trung2024reft}
Luong Trung, Xinbo Zhang, Zhanming Jie, Peng Sun, Xiaoran Jin, and Hang Li.
\newblock Reft: Reasoning with reinforced fine-tuning.
\newblock In {\em Proceedings of the 62nd Annual Meeting of the Association for Computational Linguistics (Volume 1: Long Papers)}, pages 7601--7614, 2024.

\bibitem{valmeekam2024llms}
Karthik Valmeekam, Kaya Stechly, and Subbarao Kambhampati.
\newblock Llms still can't plan; can lrms? a preliminary evaluation of openai's o1 on planbench.
\newblock {\em arXiv preprint arXiv:2409.13373}, 2024.

\bibitem{wang2024openr}
Jun Wang, Meng Fang, Ziyu Wan, Muning Wen, Jiachen Zhu, Anjie Liu, Ziqin Gong, Yan Song, Lei Chen, Lionel~M Ni, et~al.
\newblock Openr: An open source framework for advanced reasoning with large language models.
\newblock {\em arXiv preprint arXiv:2410.09671}, 2024.

\bibitem{wang2024planning}
Kevin Wang, Junbo Li, Neel~P Bhatt, Yihan Xi, Qiang Liu, Ufuk Topcu, and Zhangyang Wang.
\newblock On the planning abilities of openai's o1 models: Feasibility, optimality, and generalizability.
\newblock {\em arXiv preprint arXiv:2409.19924}, 2024.

\bibitem{wang2023plan}
Lei Wang, Wanyu Xu, Yihuai Lan, Zhiqiang Hu, Yunshi Lan, Roy Ka-Wei Lee, and Ee-Peng Lim.
\newblock Plan-and-solve prompting: Improving zero-shot chain-of-thought reasoning by large language models.
\newblock {\em arXiv preprint arXiv:2305.04091}, 2023.

\bibitem{wang2024math}
Peiyi Wang, Lei Li, Zhihong Shao, Runxin Xu, Damai Dai, Yifei Li, Deli Chen, Yu~Wu, and Zhifang Sui.
\newblock Math-shepherd: Verify and reinforce llms step-by-step without human annotations.
\newblock In {\em Proceedings of the 62nd Annual Meeting of the Association for Computational Linguistics (Volume 1: Long Papers)}, pages 9426--9439, 2024.

\bibitem{wang2023math}
Peiyi Wang, Lei Li, Zhihong Shao, RX~Xu, Damai Dai, Yifei Li, Deli Chen, Y~Wu, and Zhifang Sui.
\newblock Math-shepherd: A label-free step-by-step verifier for llms in mathematical reasoning.
\newblock {\em arXiv preprint arXiv:2312.08935}, 2023.

\bibitem{wang2024cpl}
Tianlong Wang, Xueting Han, and Jing Bai.
\newblock Cpl: Critical planning step learning boosts llm generalization in reasoning tasks.
\newblock {\em arXiv preprint arXiv:2409.08642}, 2024.

\bibitem{wang2024scibenchevaluatingcollegelevelscientific}
Xiaoxuan Wang, Ziniu Hu, Pan Lu, Yanqiao Zhu, Jieyu Zhang, Satyen Subramaniam, Arjun~R. Loomba, Shichang Zhang, Yizhou Sun, and Wei Wang.
\newblock Scibench: Evaluating college-level scientific problem-solving abilities of large language models, 2024.

\bibitem{wang2024human}
Xinru Wang, Hannah Kim, Sajjadur Rahman, Kushan Mitra, and Zhengjie Miao.
\newblock Human-llm collaborative annotation through effective verification of llm labels.
\newblock In {\em Proceedings of the CHI Conference on Human Factors in Computing Systems}, pages 1--21, 2024.

\bibitem{wang2022self}
Yizhong Wang, Yeganeh Kordi, Swaroop Mishra, Alisa Liu, Noah~A Smith, Daniel Khashabi, and Hannaneh Hajishirzi.
\newblock Self-instruct: Aligning language models with self-generated instructions.
\newblock {\em arXiv preprint arXiv:2212.10560}, 2022.

\bibitem{wang2024mmlu}
Yubo Wang, Xueguang Ma, Ge~Zhang, Yuansheng Ni, Abhranil Chandra, Shiguang Guo, Weiming Ren, Aaran Arulraj, Xuan He, Ziyan Jiang, et~al.
\newblock Mmlu-pro: A more robust and challenging multi-task language understanding benchmark.
\newblock {\em arXiv preprint arXiv:2406.01574}, 2024.

\bibitem{wang2022execution}
Zhiruo Wang, Shuyan Zhou, Daniel Fried, and Graham Neubig.
\newblock Execution-based evaluation for open-domain code generation.
\newblock {\em arXiv preprint arXiv:2212.10481}, 2022.

\bibitem{wang2024multi}
Zihan Wang, Yunxuan Li, Yuexin Wu, Liangchen Luo, Le~Hou, Hongkun Yu, and Jingbo Shang.
\newblock Multi-step problem solving through a verifier: An empirical analysis on model-induced process supervision.
\newblock {\em arXiv preprint arXiv:2402.02658}, 2024.

\bibitem{webb2023emergent}
Taylor Webb, Keith~J Holyoak, and Hongjing Lu.
\newblock Emergent analogical reasoning in large language models.
\newblock {\em Nature Human Behaviour}, 7(9):1526--1541, 2023.

\bibitem{wei2021finetuned}
Jason Wei, Maarten Bosma, Vincent~Y Zhao, Kelvin Guu, Adams~Wei Yu, Brian Lester, Nan Du, Andrew~M Dai, and Quoc~V Le.
\newblock Finetuned language models are zero-shot learners.
\newblock {\em arXiv preprint arXiv:2109.01652}, 2021.

\bibitem{wei2022emergent}
Jason Wei, Yi~Tay, Rishi Bommasani, Colin Raffel, Barret Zoph, Sebastian Borgeaud, Dani Yogatama, Maarten Bosma, Denny Zhou, Donald Metzler, et~al.
\newblock Emergent abilities of large language models.
\newblock {\em arXiv preprint arXiv:2206.07682}, 2022.

\bibitem{wei2022chain}
Jason Wei, Xuezhi Wang, Dale Schuurmans, Maarten Bosma, Fei Xia, Ed~Chi, Quoc~V Le, Denny Zhou, et~al.
\newblock Chain-of-thought prompting elicits reasoning in large language models.
\newblock {\em Advances in neural information processing systems}, 35:24824--24837, 2022.

\bibitem{wei2023simple}
Jerry Wei, Da~Huang, Yifeng Lu, Denny Zhou, and Quoc~V Le.
\newblock Simple synthetic data reduces sycophancy in large language models.
\newblock {\em arXiv preprint arXiv:2308.03958}, 2023.

\bibitem{wysocki2024llm}
Oskar Wysocki, Magdalena Wysocka, Danilo Carvalho, Alex~Teodor Bogatu, Danilo~Miranda Gusicuma, Maxime Delmas, Harriet Unsworth, and Andre Freitas.
\newblock An llm-based knowledge synthesis and scientific reasoning framework for biomedical discovery.
\newblock {\em arXiv preprint arXiv:2406.18626}, 2024.

\bibitem{xi2023rise}
Zhiheng Xi, Wenxiang Chen, Xin Guo, Wei He, Yiwen Ding, Boyang Hong, Ming Zhang, Junzhe Wang, Senjie Jin, Enyu Zhou, et~al.
\newblock The rise and potential of large language model based agents: A survey.
\newblock {\em arXiv preprint arXiv:2309.07864}, 2023.

\bibitem{xi2024agentgym}
Zhiheng Xi, Yiwen Ding, Wenxiang Chen, Boyang Hong, Honglin Guo, Junzhe Wang, Dingwen Yang, Chenyang Liao, Xin Guo, Wei He, et~al.
\newblock Agentgym: Evolving large language model-based agents across diverse environments.
\newblock {\em arXiv preprint arXiv:2406.04151}, 2024.

\bibitem{xie2024monte}
Yuxi Xie, Anirudh Goyal, Wenyue Zheng, Min-Yen Kan, Timothy~P Lillicrap, Kenji Kawaguchi, and Michael Shieh.
\newblock Monte carlo tree search boosts reasoning via iterative preference learning.
\newblock {\em arXiv preprint arXiv:2405.00451}, 2024.

\bibitem{xiong_trigo_2023}
Jing Xiong, Jianhao Shen, Ye~Yuan, Haiming Wang, Yichun Yin, Zhengying Liu, Lin Li, Zhijiang Guo, Qingxing Cao, Yinya Huang, Chuanyang Zheng, Xiaodan Liang, Ming Zhang, and Qun Liu.
\newblock {TRIGO}: {Benchmarking} {Formal} {Mathematical} {Proof} {Reduction} for {Generative} {Language} {Models}, October 2023.
\newblock arXiv:2310.10180 [cs].

\bibitem{xu2023wizardlm}
Can Xu, Qingfeng Sun, Kai Zheng, Xiubo Geng, Pu~Zhao, Jiazhan Feng, Chongyang Tao, and Daxin Jiang.
\newblock Wizardlm: Empowering large language models to follow complex instructions.
\newblock {\em arXiv preprint arXiv:2304.12244}, 2023.

\bibitem{yang2024qwen2}
An~Yang, Baosong Yang, Beichen Zhang, Binyuan Hui, Bo~Zheng, Bowen Yu, Chengyuan Li, Dayiheng Liu, Fei Huang, Haoran Wei, et~al.
\newblock Qwen2. 5 technical report.
\newblock {\em arXiv preprint arXiv:2412.15115}, 2024.

\bibitem{yang2024large}
Chengrun Yang, Xuezhi Wang, Yifeng Lu, Hanxiao Liu, Quoc~V Le, Denny Zhou, and Xinyun Chen.
\newblock Large language models as optimizers.
\newblock In {\em The Twelfth International Conference on Learning Representations}, 2024.

\bibitem{yang2023leandojotheoremprovingretrievalaugmented}
Kaiyu Yang, Aidan~M. Swope, Alex Gu, Rahul Chalamala, Peiyang Song, Shixing Yu, Saad Godil, Ryan Prenger, and Anima Anandkumar.
\newblock Leandojo: Theorem proving with retrieval-augmented language models, 2023.

\bibitem{yao2022webshop}
Shunyu Yao, Howard Chen, John Yang, and Karthik Narasimhan.
\newblock Webshop: Towards scalable real-world web interaction with grounded language agents.
\newblock {\em Advances in Neural Information Processing Systems}, 35:20744--20757, 2022.

\bibitem{yao2024tree}
Shunyu Yao, Dian Yu, Jeffrey Zhao, Izhak Shafran, Tom Griffiths, Yuan Cao, and Karthik Narasimhan.
\newblock Tree of thoughts: Deliberate problem solving with large language models.
\newblock {\em Advances in Neural Information Processing Systems}, 36, 2024.

\bibitem{yao2022react}
Shunyu Yao, Jeffrey Zhao, Dian Yu, Nan Du, Izhak Shafran, Karthik Narasimhan, and Yuan Cao.
\newblock React: Synergizing reasoning and acting in language models.
\newblock {\em arXiv preprint arXiv:2210.03629}, 2022.

\bibitem{ye2024reevo}
Haoran Ye, Jiarui Wang, Zhiguang Cao, Federico Berto, Chuanbo Hua, Haeyeon Kim, Jinkyoo Park, and Guojie Song.
\newblock Reevo: Large language models as hyper-heuristics with reflective evolution.
\newblock {\em arXiv preprint arXiv:2402.01145}, 2024.

\bibitem{young2022abductionrulestrainingtransformersexplain}
Nathan Young, Qiming Bao, Joshua Bensemann, and Michael Witbrock.
\newblock Abductionrules: Training transformers to explain unexpected inputs, 2022.

\bibitem{yu2024fincon}
Yangyang Yu, Zhiyuan Yao, Haohang Li, Zhiyang Deng, Yupeng Cao, Zhi Chen, Jordan~W Suchow, Rong Liu, Zhenyu Cui, Zhaozhuo Xu, et~al.
\newblock Fincon: A synthesized llm multi-agent system with conceptual verbal reinforcement for enhanced financial decision making.
\newblock {\em arXiv preprint arXiv:2407.06567}, 2024.

\bibitem{yuan2023largelanguagemodelsperform}
Zheng Yuan, Hongyi Yuan, Chuanqi Tan, Wei Wang, and Songfang Huang.
\newblock How well do large language models perform in arithmetic tasks?, 2023.

\bibitem{zellers2018swag}
Rowan Zellers, Yonatan Bisk, Roy Schwartz, and Yejin Choi.
\newblock Swag: A large-scale adversarial dataset for grounded commonsense inference.
\newblock In {\em Proceedings of the 2018 Conference on Empirical Methods in Natural Language Processing}, pages 93--104, 2018.

\bibitem{zellers2019hellaswag}
Rowan Zellers, Ari Holtzman, Yonatan Bisk, Ali Farhadi, and Yejin Choi.
\newblock Hellaswag: Can a machine really finish your sentence?
\newblock In {\em Proceedings of the 57th Annual Meeting of the Association for Computational Linguistics}, pages 4791--4800, 2019.

\bibitem{zeng2024perceive}
Qingbin Zeng, Qinglong Yang, Shunan Dong, Heming Du, Liang Zheng, Fengli Xu, and Yong Li.
\newblock Perceive, reflect, and plan: Designing llm agent for goal-directed city navigation without instructions.
\newblock {\em arXiv preprint arXiv:2408.04168}, 2024.

\bibitem{zeng2024llmbar}
Zhiyuan Zeng, Jiatong Yu, Tianyu Gao, Yu~Meng, Tanya Goyal, and Danqi Chen.
\newblock Evaluating large language models at evaluating instruction following.
\newblock In {\em International Conference on Learning Representations (ICLR)}, 2024.

\bibitem{zhang2024good}
Chenhui Zhang and Sherrie Wang.
\newblock Good at captioning, bad at counting: Benchmarking gpt-4v on earth observation data.
\newblock {\em arXiv preprint arXiv:2401.17600}, 2024.

\bibitem{zhang2024rest}
Dan Zhang, Sining Zhoubian, Ziniu Hu, Yisong Yue, Yuxiao Dong, and Jie Tang.
\newblock Rest-mcts*: Llm self-training via process reward guided tree search.
\newblock {\em arXiv preprint arXiv:2406.03816}, 2024.

\bibitem{zhang2024large}
Danyang Zhang, Lu~Chen, Situo Zhang, Hongshen Xu, Zihan Zhao, and Kai Yu.
\newblock Large language models are semi-parametric reinforcement learning agents.
\newblock {\em Advances in Neural Information Processing Systems}, 36, 2024.

\bibitem{zhang2024llama}
Di~Zhang, Jianbo Wu, Jingdi Lei, Tong Che, Jiatong Li, Tong Xie, Xiaoshui Huang, Shufei Zhang, Marco Pavone, Yuqiang Li, et~al.
\newblock Llama-berry: Pairwise optimization for o1-like olympiad-level mathematical reasoning.
\newblock {\em arXiv preprint arXiv:2410.02884}, 2024.

\bibitem{zhang2024aflow}
Jiayi Zhang, Jinyu Xiang, Zhaoyang Yu, Fengwei Teng, Xionghui Chen, Jiaqi Chen, Mingchen Zhuge, Xin Cheng, Sirui Hong, Jinlin Wang, et~al.
\newblock Aflow: Automating agentic workflow generation.
\newblock {\em arXiv preprint arXiv:2410.10762}, 2024.

\bibitem{zhang2023instruction}
Shengyu Zhang, Linfeng Dong, Xiaoya Li, Sen Zhang, Xiaofei Sun, Shuhe Wang, Jiwei Li, Runyi Hu, Tianwei Zhang, Fei Wu, et~al.
\newblock Instruction tuning for large language models: A survey.
\newblock {\em arXiv preprint arXiv:2308.10792}, 2023.

\bibitem{zhang2024agent}
Wenqi Zhang, Ke~Tang, Hai Wu, Mengna Wang, Yongliang Shen, Guiyang Hou, Zeqi Tan, Peng Li, Yueting Zhuang, and Weiming Lu.
\newblock Agent-pro: Learning to evolve via policy-level reflection and optimization.
\newblock {\em arXiv preprint arXiv:2402.17574}, 2024.

\bibitem{zhang2022automatic}
Zhuosheng Zhang, Aston Zhang, Mu~Li, and Alex Smola.
\newblock Automatic chain of thought prompting in large language models.
\newblock {\em arXiv preprint arXiv:2210.03493}, 2022.

\bibitem{zhao2024expel}
Andrew Zhao, Daniel Huang, Quentin Xu, Matthieu Lin, Yong-Jin Liu, and Gao Huang.
\newblock Expel: Llm agents are experiential learners.
\newblock In {\em Proceedings of the AAAI Conference on Artificial Intelligence}, volume~38, pages 19632--19642, 2024.

\bibitem{zhao2023survey}
Wayne~Xin Zhao, Kun Zhou, Junyi Li, Tianyi Tang, Xiaolei Wang, Yupeng Hou, Yingqian Min, Beichen Zhang, Junjie Zhang, Zican Dong, et~al.
\newblock A survey of large language models.
\newblock {\em arXiv preprint arXiv:2303.18223}, 2023.

\bibitem{zhao2022multihierttnumericalreasoningmulti}
Yilun Zhao, Yunxiang Li, Chenying Li, and Rui Zhang.
\newblock Multihiertt: Numerical reasoning over multi hierarchical tabular and textual data, 2022.

\bibitem{zheng2022minif2fcrosssystembenchmarkformal}
Kunhao Zheng, Jesse~Michael Han, and Stanislas Polu.
\newblock Minif2f: a cross-system benchmark for formal olympiad-level mathematics, 2022.

\bibitem{zhong2024evaluation}
Tianyang Zhong, Zhengliang Liu, Yi~Pan, Yutong Zhang, Yifan Zhou, Shizhe Liang, Zihao Wu, Yanjun Lyu, Peng Shu, Xiaowei Yu, et~al.
\newblock Evaluation of openai o1: Opportunities and challenges of agi.
\newblock {\em arXiv preprint arXiv:2409.18486}, 2024.

\bibitem{zhou2024lima}
Chunting Zhou, Pengfei Liu, Puxin Xu, Srinivasan Iyer, Jiao Sun, Yuning Mao, Xuezhe Ma, Avia Efrat, Ping Yu, Lili Yu, et~al.
\newblock Lima: Less is more for alignment.
\newblock {\em Advances in Neural Information Processing Systems}, 36, 2024.

\bibitem{zhou2022least}
Denny Zhou, Nathanael Sch{\"a}rli, Le~Hou, Jason Wei, Nathan Scales, Xuezhi Wang, Dale Schuurmans, Claire Cui, Olivier Bousquet, Quoc Le, et~al.
\newblock Least-to-most prompting enables complex reasoning in large language models.
\newblock {\em arXiv preprint arXiv:2205.10625}, 2022.

\bibitem{zhou2023webarena}
Shuyan Zhou, Frank~F Xu, Hao Zhu, Xuhui Zhou, Robert Lo, Abishek Sridhar, Xianyi Cheng, Tianyue Ou, Yonatan Bisk, Daniel Fried, et~al.
\newblock Webarena: A realistic web environment for building autonomous agents.
\newblock {\em arXiv preprint arXiv:2307.13854}, 2023.

\bibitem{zhou2024archer}
Yifei Zhou, Andrea Zanette, Jiayi Pan, Sergey Levine, and Aviral Kumar.
\newblock Archer: Training language model agents via hierarchical multi-turn rl.
\newblock {\em arXiv preprint arXiv:2402.19446}, 2024.

\bibitem{zhou2024synergizing}
Zhilun Zhou, Jingyang Fan, Yu~Liu, Fengli Xu, Depeng Jin, and Yong Li.
\newblock Synergizing llm agents and knowledge graph for socioeconomic prediction in lbsn.
\newblock {\em arXiv preprint arXiv:2411.00028}, 2024.

\bibitem{zhugegptswarm}
Mingchen Zhuge, Wenyi Wang, Louis Kirsch, Francesco Faccio, Dmitrii Khizbullin, and J{\"u}rgen Schmidhuber.
\newblock Gptswarm: Language agents as optimizable graphs.
\newblock In {\em Forty-first International Conference on Machine Learning}.

\end{thebibliography}

\end{document}